\documentclass[journal]{IEEEtran}
\usepackage{hyperref}
\usepackage{amsmath,amsfonts}
\usepackage{array}
\usepackage[caption=false,font=normalsize,labelfont=sf,textfont=sf]{subfig}
\usepackage{textcomp}
\usepackage{stfloats}
\usepackage{url}
\usepackage{verbatim}
\usepackage{graphicx}
\usepackage{multirow}
\usepackage{makecell}
\usepackage{algorithm}
\usepackage{algpseudocode}
\usepackage{cleveref}
\usepackage{svg}
\usepackage{threeparttable} 

\usepackage{tikz,xcolor}

\hypersetup{hidelinks,
	colorlinks=true,
	allcolors=black,
	pdfstartview=Fit,
	breaklinks=true}

\definecolor{lime}{HTML}{A6CE39}
\DeclareRobustCommand{\orcidicon}{
\begin{tikzpicture}
\draw[lime, fill=lime] (0,0)
circle[radius=0.16]
node[white]{{\fontfamily{qag}\selectfont \tiny \.{I}D}};
\end{tikzpicture}
\hspace{-2mm}
}
\foreach \x in {A, ..., Z}{%
\expandafter\xdef\csname orcid\x\endcsname{\noexpand\href{https://orcid.org/\csname orcidauthor\x\endcsname}{\noexpand\orcidicon}}
}

\def\BibTeX{{\rm B\kern-.05em{\sc i\kern-.025em b}\kern-.08em
    T\kern-.1667em\lower.7ex\hbox{E}\kern-.125emX}}
\usepackage{balance}
\begin{document} 
\title{Genetic Programming with Reinforcement Learning Trained Transformer for Real-World Dynamic Scheduling Problems}
\author{Xinan Chen\hspace{-1.5mm}\orcidA{}, \IEEEmembership{Member, IEEE, }
        Rong~Qu\hspace{-1.5mm}\orcidC{}, \IEEEmembership{Senior Member, IEEE, }
        Jing~Dong\hspace{-1.5mm}\orcidD{},
        Ruibin~Bai\hspace{-1.5mm}\orcidB{}, \IEEEmembership{Senior Member, IEEE, }
        and Yaochu Jin \hspace{-1.5mm}\orcidE{}, \IEEEmembership{Fellow, IEEE}
\thanks{This work is supported by the National Natural Science Foundation of China (Grant No.72071116)  and Ningbo Municipal Bureau of Science and Technology (Grant No. 2023Z237). (Corresponding author: Ruibin Bai.)}

\thanks{Xinan Chen and Ruibin Bai is with the Digital Port Technologies Lab, School of Computer Science, University of Nottingham Ningbo China, Ningbo 315100, China (email: xinan.chen@nottingham.edu.cn, ruibin.bai@nottingham.edu.cn).}
\thanks{Rong Qu is with the School of Computer Science, University of Nottingham, Nottingham NG72RD, UK (email: rong.qu@nottingham.ac.uk).}
\thanks{Jing Dong is with the Department of Engineering, University of Cambridge, Cambridge CB21TN, UK (email: djjlyxfo@gmail.com).}
\thanks{Yaochu Jin is with the School of Engineering, Westlake University,
Hangzhou 310030, China (email: jinyaochu@westlake.edu.cn).}

}

\markboth{Journal of \LaTeX\ Class Files,~Vol.~18, No.~9, September~2020}%
{How to Use the IEEEtran \LaTeX \ Templates}

\maketitle

\begin{abstract}
Dynamic scheduling in real-world environments often struggles to adapt to unforeseen disruptions, making traditional static scheduling methods and human-designed heuristics inadequate. This paper introduces an innovative approach that combines Genetic Programming (GP) with a Transformer trained through Reinforcement Learning (GPRT), specifically designed to tackle the complexities of dynamic scheduling scenarios.
GPRT leverages the Transformer to refine heuristics generated by GP while also seeding and guiding the evolution of GP. This dual functionality enhances the adaptability and effectiveness of the scheduling heuristics, enabling them to better respond to the dynamic nature of real-world tasks. The efficacy of this integrated approach is demonstrated through a practical application in container terminal truck scheduling, where the GPRT method outperforms traditional GP, standalone Transformer methods, and other state-of-the-art competitors.
The key contribution of this research is the development of the GPRT method, which showcases a novel combination of GP and Reinforcement Learning (RL) to produce robust and efficient scheduling solutions. Importantly, GPRT is not limited to container port truck scheduling; it offers a versatile framework applicable to various dynamic scheduling challenges. Its practicality, coupled with its interpretability and ease of modification, makes it a valuable tool for diverse real-world scenarios.
\end{abstract}

\begin{IEEEkeywords}
reinforcement learning, transformer, genetic programming, dynamic scheduling, truck scheduling

\end{IEEEkeywords}



\section{Introduction}

Most real-world scheduling problems exist within dynamic environments, where unpredictable real-time events such as unforeseen machine failures, the arrival of urgent jobs, due date alterations, and unexpected weather changes are not just possible but often inevitable. These events can disrupt scheduled plans, rendering them unfeasible when executed. There is a notable disconnect between theoretical scheduling models and their practical application, primarily because classical scheduling theories struggle to adapt to the dynamic nature of real-world environments \cite{ouelhadj2009survey}. In response to this disparity, recent research trends in scheduling have increasingly focused on developing theories that are more relevant and applicable to real-world scenarios.

Dynamic scheduling has consequently received increasing attention and focus in recent years. This growing interest stems from the recognition that real-world dynamic scheduling problems significantly diverge from conventional scheduling methods. The challenge lies in finding effective scheduling solutions that can adapt to these dynamic contexts \cite{bukkur2018review}. The complexity and unpredictability of real-world environments demand flexible, adaptable scheduling approaches that respond to real-time, unexpected changes. Thus, the field of dynamic scheduling is evolving to bridge the gap between theory and practice, offering solutions that are both theoretically sound and practically feasible in the ever-changing landscape of real-world operations.

Due to the constraints inherent in traditional static scheduling methods, dynamic scheduling problems in real-world production environments are typically addressed using simple decision trees or heuristics. This methodology is widely favored across various enterprises for its rapid decision-making capabilities, transparency, practicality, and dependability \cite{edelkamp2011heuristic}. However, these simple, expert-designed heuristics have inherent limitations. While effective in straightforward operational environments, their efficiency diminishes due to contemporary production settings' complex and evolving dynamics. As a result, more sophisticated machine learning-based methods, such as Reinforcement Learning (RL) and Genetic Programming (GP), are increasingly being adopted for dynamic scheduling. Both approaches extend beyond conventional solution spaces, venturing into the broader hyperspace for solution generation. In the scheduling process, these trained algorithms consider the current environmental context and the influence of various uncertainties, thereby generating the most apt solutions for specific scenarios. Such machine learning-based methodologies have markedly improved the efficiency of dynamic scheduling in complex real-world operational environments, effectively navigating the challenges inherent in these contexts.

Both GP and RL offer distinct advantages in addressing dynamic scheduling issues. GP is adept at swiftly identifying feasible solutions, while RL-based methods, despite potentially higher training costs, often demonstrate superior performance and efficiency compared to GP-based approaches. However, traditional reinforcement learning faces challenges in complex, large-scale, multi-action, real-world dynamic scheduling scenarios, primarily due to an excessive action space and sparse rewards, leading to difficulties in achieving convergence. Conversely, while GP is effective at quickly finding feasible solutions, it struggles with further optimizing solution quality and deriving alternative solutions from feasible ones. This challenge arises from the intrinsic uncertainties in GP's search process, where evolutionary search algorithms can quickly identify a satisfactory solution. However, advancing beyond this to improve solution quality often requires random crossover or mutation, thus incurring significant costs for enhanced solutions. In contrast, RL expedites the optimization of solution quality through a guided search process involving rewards and backpropagation of errors, enabling faster identification of superior solutions. Consequently, this paper introduces a novel approach, Genetic Programming with Reinforcement Learning Trained Transformer (GPRT), which combines the strengths of both GP and RL. This method not only identifies solutions rapidly but also significantly improves their quality.

In the GPRT framework, GP and the Transformer utilize the same expression format to represent solutions. This shared format allows the Transformer to refine GP's solutions, enhancing outcomes. Additionally, the Transformer can generate a series of new individuals to overcome the limited population diversity faced during GP's evolutionary process, further augmenting GP's performance. Moreover, data generated during GP's training phase can be leveraged to expedite the training of the Transformer, substantially reducing associated training costs. Furthermore, GPRT addresses the challenges posed by the traditional RL's 'black box' nature, which is often challenging to interpret, modify, and comprehend, resulting in its limited application in real-world production environments. Collaborative insights from our work with Ningbo-Zhoushan Port Company Limited (the world's largest container port) indicate a preference for the understandable and modifiable expressions generated by GP compared to RL-based methods. This preference was a key motivator in developing the GPRT method, aiming to enhance the performance of traditional GP. 
The primary contributions of this paper are as follows:

\begin{itemize}
    \item \textbf{Innovative Hybrid Framework}: This paper introduces a novel integration of GP and RL through the GPRR and GPRT frameworks, specifically tailored for dynamic scheduling problems, significantly advancing the state of the art in optimization techniques.

    \item \textbf{Enhanced Heuristic Performance}: By combining the strengths of GP and RL, we demonstrate that the proposed methods effectively improve the adaptability and efficiency of heuristics, allowing for better performance in complex and uncertain environments.

    \item \textbf{Interpretability and Complexity Reduction}: The integration of RNNs and Transformers with GP not only enhances performance but also results in smaller, more interpretable individuals, addressing the challenges of complexity and interpretability often associated with traditional optimization methods.

    \item \textbf{Real-World Applicability}: Empirical results from applying the GPRT framework to container terminal truck scheduling highlight its practical effectiveness, showcasing the framework's ability to outperform existing approaches in real-world scenarios while providing actionable insights for decision-making.
\end{itemize}

The rest of this paper is organized as follows. Section \ref{section:back} reviews related work and provides background information on the real-world dynamic scheduling problem. Section \ref{section:problem} introduces and formulates the specific dynamic truck scheduling problem under consideration. The proposed GPRT method is delineated in Section \ref{section:method}. Section \ref{section:experiment} delineates the experimental outcomes and provides ablation \& sensitivity analysis of proposed methods.
Finally, conclusions are drawn in Section \ref{section:conclusion}.

\section{Background}\label{section:back}
In this section, we provide the foundational concepts and background that underpin our proposed approach. We begin by exploring the characteristics and challenges of \emph{real-world dynamic scheduling}, highlighting the complexities that arise in practical applications. We then delve into \emph{Genetic Programming}, discussing its principles and how it has been applied to scheduling problems. Following this, we examine \emph{Reinforcement Learning}, focusing on its ability to learn adaptive policies in dynamic environments. Lastly, we introduce the \emph{Transformer} architecture, emphasizing its strengths in sequence modelling and potential benefits for scheduling tasks.

\subsection{Real-World Dynamic Scheduling}
Scheduling is fundamental in operations management, involving the allocation of limited resources to tasks over time to optimize objectives such as minimizing completion time or maximizing resource utilization \cite{schumann2022scheduling}. In real-world environments, scheduling problems are inherently \emph{dynamic} due to uncertainties and unforeseen events that disrupt planned schedules \cite{qiao2018novel}. These problems require continual adaptation to changes like unpredictable job arrivals, variable processing times, resource unavailability, and environmental factors.

Dynamic scheduling is crucial because static methods, which assume complete information and unchanging conditions, often fail to cope with the variability and uncertainty present in practical applications \cite{begg2014uncertainty}. Effective dynamic scheduling enhances responsiveness and efficiency by allowing systems to adjust in real-time to disruptions, thereby improving robustness and maintaining performance levels \cite{lu2017hybrid}.

A significant challenge in addressing real-world dynamic scheduling is the difference between real-world and experimental data. Real-world data is often noisy, incomplete, and derived from complex, non-stationary distributions \cite{dixit2023contemporary}. It exhibits high variability due to factors like fluctuating demand and environmental conditions. In contrast, experimental data is typically generated under controlled conditions with simplifying assumptions, such as deterministic processing times and known probability distributions \cite{maxwell2017designing}.

These discrepancies mean that models developed on experimental data may not generalize well to real-world scenarios, leading to performance degradation when faced with uncertainties not present in controlled data \cite{walker2003defining}. Therefore, robust methods capable of handling the complexities of real-world data, including noise and incomplete information, are necessary \cite{jin2005evolutionary}.

To address the complexities of real-world dynamic scheduling, advanced methodologies have been proposed. Traditional optimization techniques like mixed-integer programming \cite{fatemi2023scheduling,deng2018optimal} and heuristic algorithms \cite{thilagavathi2014survey} often struggle with scalability and adaptability in dynamic environments. Metaheuristic approaches such as genetic algorithms \cite{zhu2020making}, tabu search \cite{firme2023agent}, and simulated annealing \cite{chou2008simulated} offer more flexibility but may not effectively handle the stochastic nature of real-world data. They can find reasonable solutions but cannot accommodate uncertainty, meaning the solution must be recalculated after the environment changes \cite{renke2021review}.

Recently, adaptive methods that learn over time have gained prominence. Among these, GP and RL have become particularly popular due to their ability to handle complex, uncertain environments~\cite{xu2024genetic}. RL learns optimal policies through interaction with the environment, making it suitable for sequential decision-making under uncertainty~\cite{shyalika2020reinforcement}. GP, on the other hand, evolves heuristics that adjust to changing conditions, providing flexible and interpretable solutions~\cite{zhang2021genetic}. Their popularity stems from their capacity to improve over time, handle non-linearities, and adapt to uncertain environments without extensive reprogramming.

\subsection{Genetic Programming}

Genetic Programming is an evolutionary computation technique that evolves computer programs or mathematical expressions to solve complex problems~\cite{koza1994genetic}. It simulates the process of natural evolution by applying genetic operators such as selection, crossover, mutation, and reproduction to a population of candidate solutions. Most commonly, GP represents programs as tree structures, where internal nodes correspond to functions or operations, and leaves represent inputs or terminals.

\begin{figure}[htbp]
\centering
\includegraphics[width=8.8cm]{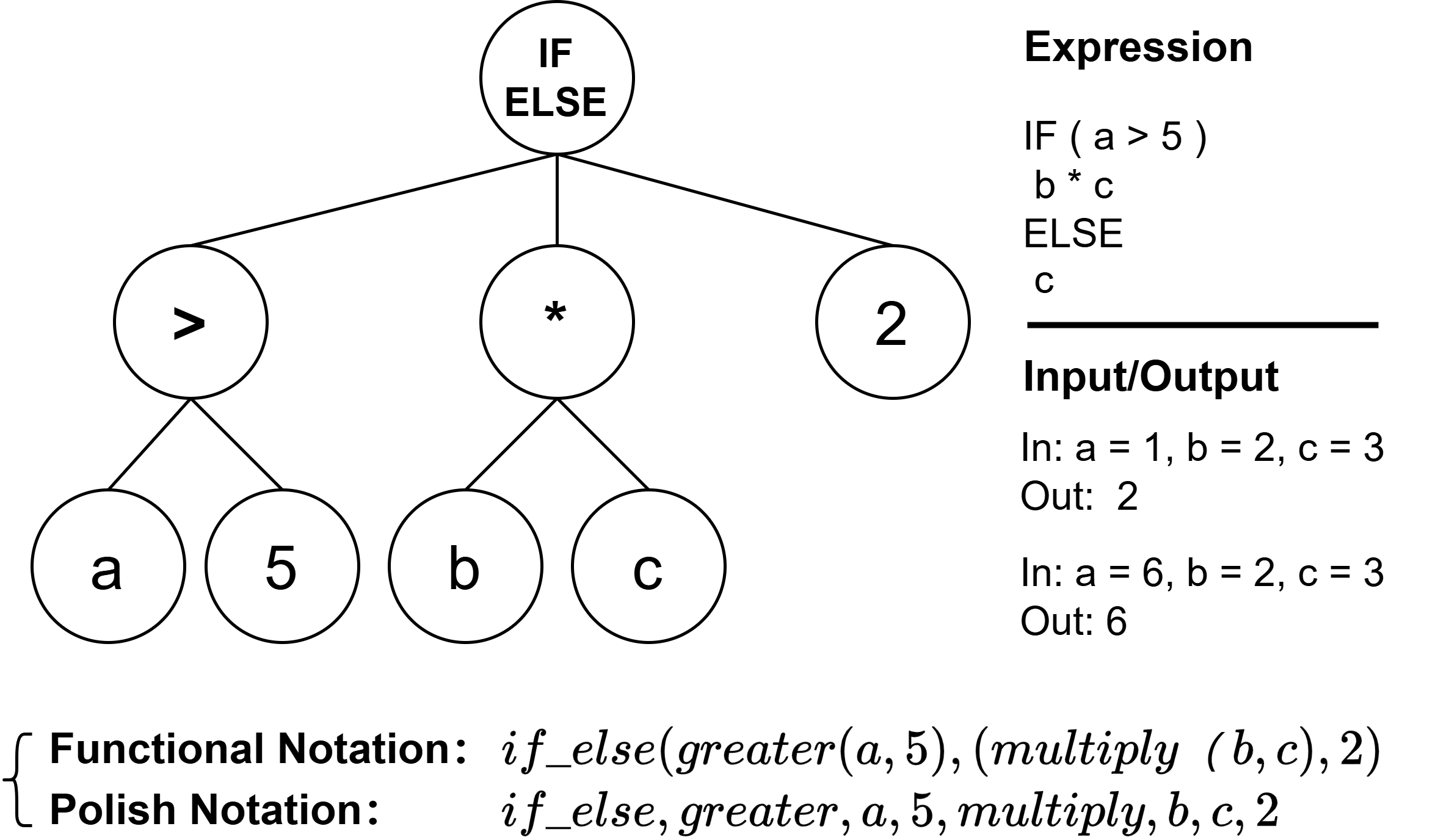}
\caption{An Example of a Genetic Programming Tree Structure}
\label{gp-exp}
\end{figure}

This paper follows the common tree-based GP representation (Fig. \ref{gp-exp}) because it naturally captures hierarchical relationships and enhances interpretability through easy visualization and translation into mathematical expressions\cite{vanneschi2014survey}. As illustrated in Fig. \ref{gp-exp}, it can produce different outputs based on varying input parameters. As the environment changes—such as fluctuations in job arrivals, processing times, or resource availability—the same GP-evolved heuristic adapts by simply receiving updated inputs, eliminating the need for retraining or reconfiguring. Once evolved, a GP individual generates scheduling decisions rapidly through straightforward computations, unlike methods requiring extensive recomputation when the environment changes. This efficiency makes GP especially suitable for real-time or near-real-time dynamic scheduling applications.

Furthermore, the tree-based GP structure uniquely facilitates integration with neural networks. Each GP-generated tree-based GP individual can be represented in Functional or Polish notation (Fig. \ref{gp-exp}), effectively treating it as a sequence of tokens or text. This compatibility allows us to utilize sequence models like recurrent neural network (RNN) \cite{medsker1999recurrent}, Long Short-Term Memory (LSTM) \cite{hochreiter1997long}, and Transformer \cite{vaswani2017attention}. By training these neural networks to generate such notations, we enable them to produce GP-like expressions.

Over the years, GP has been successfully applied to various engineering and scheduling problems, demonstrating versatility and robustness across diverse contexts~\cite{ahvanooey2019survey}. Its ability to automatically discover and evolve heuristics makes it a powerful tool for tackling complex and dynamic environments.

Compared to other machine learning and optimization methods like decision trees, logistic regression, support vector machines, and artificial neural networks, GP offers several key advantages:
\begin{itemize} 
\item \textbf{Generative Flexibility}: GP is inherently generative, capable of evolving programs that represent complex, nonlinear relationships within data~\cite{tay2008evolving}. 
\item \textbf{Powerful Search Capability}: The evolutionary mechanisms in GP facilitate the exploration of vast and complex solution spaces~\cite{banzhaf2024combinatorics}. 
\item \textbf{Interpretability and Efficiency}: The solutions generated by GP, often in the form of mathematical expressions or rule sets, are partially interpretable~\cite{riolo2010genetic}. 
\end{itemize}

Due to these advantages, GP shows significant potential not only in the context of the dynamic scheduling problem \cite{jakobovic2006dynamic, chen2020data} discussed in this study but also in a wide range of other complex optimization problems encountered in real-world scenarios like image classification \cite{bi2020genetic}, symbolic regression \cite{chen2018improving}, weather forecasting \cite{niazkar2023machine} etc. Its ability to generate adaptable, interpretable, and efficient heuristics makes it a valuable tool for addressing challenges posed by dynamic and uncertain environments.

However, despite the strong performance of GP in scheduling problems and its proven superiority over traditional expert systems \cite{sickel2010genetic}, it faces several notable limitations. Firstly, GP's training process does not inherently learn explicit knowledge about constructing scheduling strategies; its search process remains largely stochastic, lacking the accumulation of learned insights or patterns \cite{sette2001genetic}. This means that when transitioning to a different scheduling problem or when significant changes occur in the scheduling environment, GP requires complete retraining from scratch. Secondly, GP lacks effective local search capabilities. While it can quickly converge to a reasonably good solution, further refinement to find even better solutions is often challenging, hindering incremental improvements \cite{winkler2007advanced}. 

To overcome these limitations, we propose to integrate GP with RL. By combining GP's global search strengths with RL's ability to learn and adapt from interactions, we aim to enhance GP's effectiveness and adaptability in dynamic scheduling environments, enabling it to build upon accumulated knowledge and perform more efficient local searches.

\subsection{Reinforcement Learning}
\begin{figure}[htbp]
\centering
\includegraphics[width=8cm]{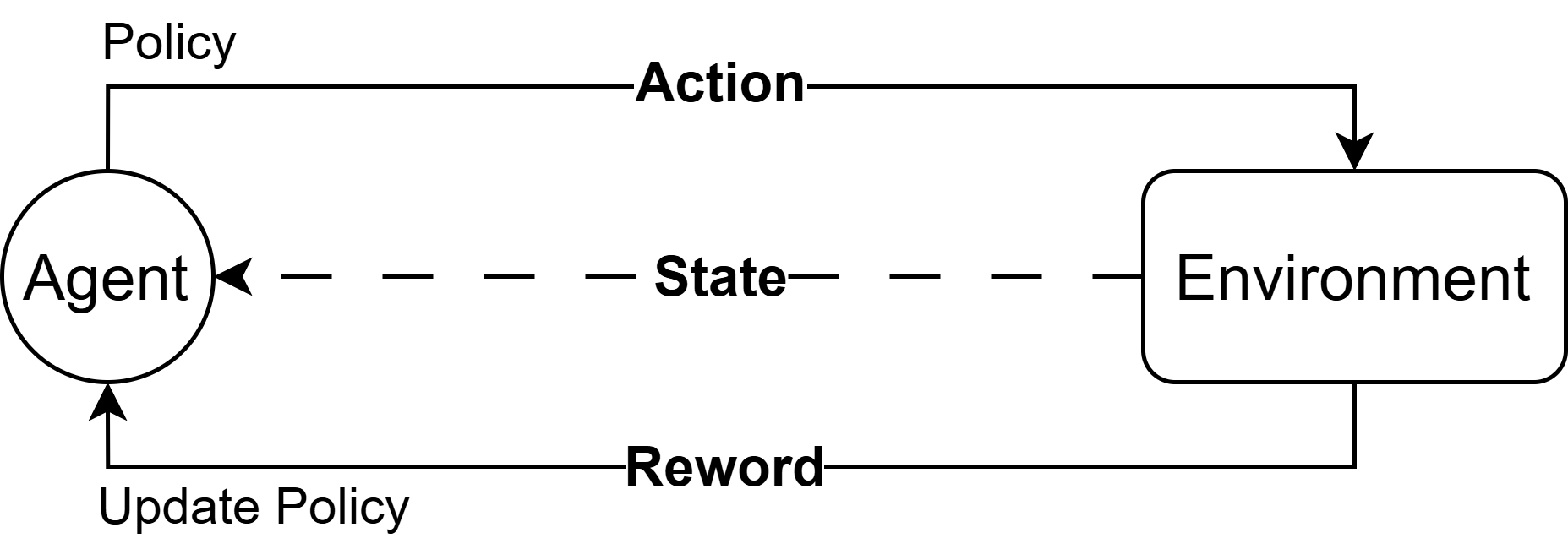}
\caption{Reinforcement Learning}
\label{rl-str}
\end{figure}

Reinforcement Learning is a branch of machine learning where agents learn to make decisions by interacting with an environment to maximize cumulative rewards \cite{kaelbling1996reinforcement}. Unlike supervised learning, RL does not require labeled input-output pairs and learns optimal behaviors through trial and error \cite{li2017deep}.
It enables agents to interact with the environment, continuously learning and refining their policies guided by rewards, as illustrated in Fig.~\ref{rl-str}.

RL has achieved remarkable success in domains that contain complexity and uncertainty. For instance, RL algorithms have attained superhuman performance in games like Go and chess, as demonstrated by AlphaGo and AlphaZero~\cite{silver2017mastering}. In robotics, RL allows robots to learn complex motor skills through interactions with their environment, facilitating tasks like manipulation and locomotion~\cite{levine2016end}. In resource management, RL has been used to optimize cloud computing resources and data center energy consumption~\cite{mao2016resource}.

In scheduling problems, RL has shown promise in handling dynamic and stochastic environments. It can adapt to changes in real-time and make sequential decisions that consider the long-term impact on system performance. RL methods have been applied to optimize job-shop scheduling~\cite{liu2023dynamic}, traffic signal control~\cite{wei2021recent}, and dynamic vehicle routing~\cite{jin2024container}, areas where traditional optimization methods often struggle due to complexity and uncertainty.

RL's strength lies in its ability to continuously refine solutions by learning from environmental interactions. As agents gather more experience, they update their policies to improve performance, adapting to environmental changes. However, RL also faces challenges, such as the need for large amounts of data to learn effective policies, potential difficulties in convergence, and the lack of interpretability of the learned policies, especially when deep neural networks are used.

Combining GP with RL offers a potential solution to these challenges. GP evolves interpretable heuristics and, during its evolutionary process, generates a large number of diverse heuristic solutions and their fitness. This abundance of data provides a rich training ground for the RL agent. Meanwhile, RL continuously refines these heuristics based on feedback from the environment. This integration leverages the strengths of both methods: GP's ability to generate flexible and interpretable solutions and produce ample data and RL's capacity to learn from experience and adapt to changes. By combining them, we aim to enhance the adaptability and performance of scheduling heuristics in dynamic environments.

One of the main challenges in integrating GP and RL is finding an effective method to enable them to work together cohesively, given their different representations and learning paradigms. To address this, we propose to utilize the Transformer architecture as a bridge between GP and RL. The Transformer, renowned for its success in natural language processing~\cite{lin2022survey}, can process sequences of tokens, making it suitable for representing GP individuals in a sequential format, such as bracket-free Polish notation. By training the Transformer using RL algorithms, we enable it to generate and refine GP-like expressions, facilitating a cohesive integration of GP and RL. This approach provides a unified representation that combines GP's interpretability with RL's learning capabilities, potentially leading to superior performance in dynamic scheduling problems.

\subsection{Transformer}
The Transformer architecture, introduced by Vaswani et al.\cite{vaswani2017attention}, is a groundbreaking approach to sequence modelling and transduction tasks that has fundamentally reshaped the field of natural language processing (NLP). Unlike traditional models that rely on recurrent or convolutional structures, the Transformer utilizes self-attention mechanisms to capture global dependencies between input and output sequences. This innovation allows for more efficient parallelization during training and has significantly reduced training times compared to recurrent neural networks\cite{zeyer2019comparison}.

Transformers have achieved remarkable success in various applications, becoming the backbone of many state-of-the-art models in NLP, such as BERT~\cite{kenton2019bert} and GPT~\cite{achiam2023gpt}. They excel in tasks like machine translation, language modelling, and text generation. Their ability to model complex patterns and generate coherent, contextually relevant sequences has made them invaluable in processing and understanding sequential data. Moreover, this capability extends beyond NLP, making Transformers suitable for a wide range of sequence modelling tasks.

Leveraging the knowledge acquired during training, Transformers can generate sequences of tokens that maximize a given reward function. In the context of dynamic scheduling tasks, this ability enables the Transformer to produce sequences expected to yield optimal scheduling performance. By learning from the environment and adjusting its outputs accordingly, the Transformer can adapt to changing conditions and generate solutions that improve over time.

By representing these sequences as bracket-free Polish notation, which can be converted into GP trees, we bridge the gap between the Transformer and GP, allowing us to integrate them effectively. This representation enables the Transformer to generate candidate GP expressions in a structured format compatible with GP's tree-based individuals. By employing RL to train the Transformer, we enable it to generate and refine tree-based GP expressions more efficiently. Unlike GP, which relies heavily on random search to explore new candidate solutions \cite{kinnear1994advances}, the Transformer can predict which candidates are more likely to yield better performance based on the knowledge acquired during training. 

\begin{figure}[htbp]
\centering
\includegraphics[width=8.8cm]{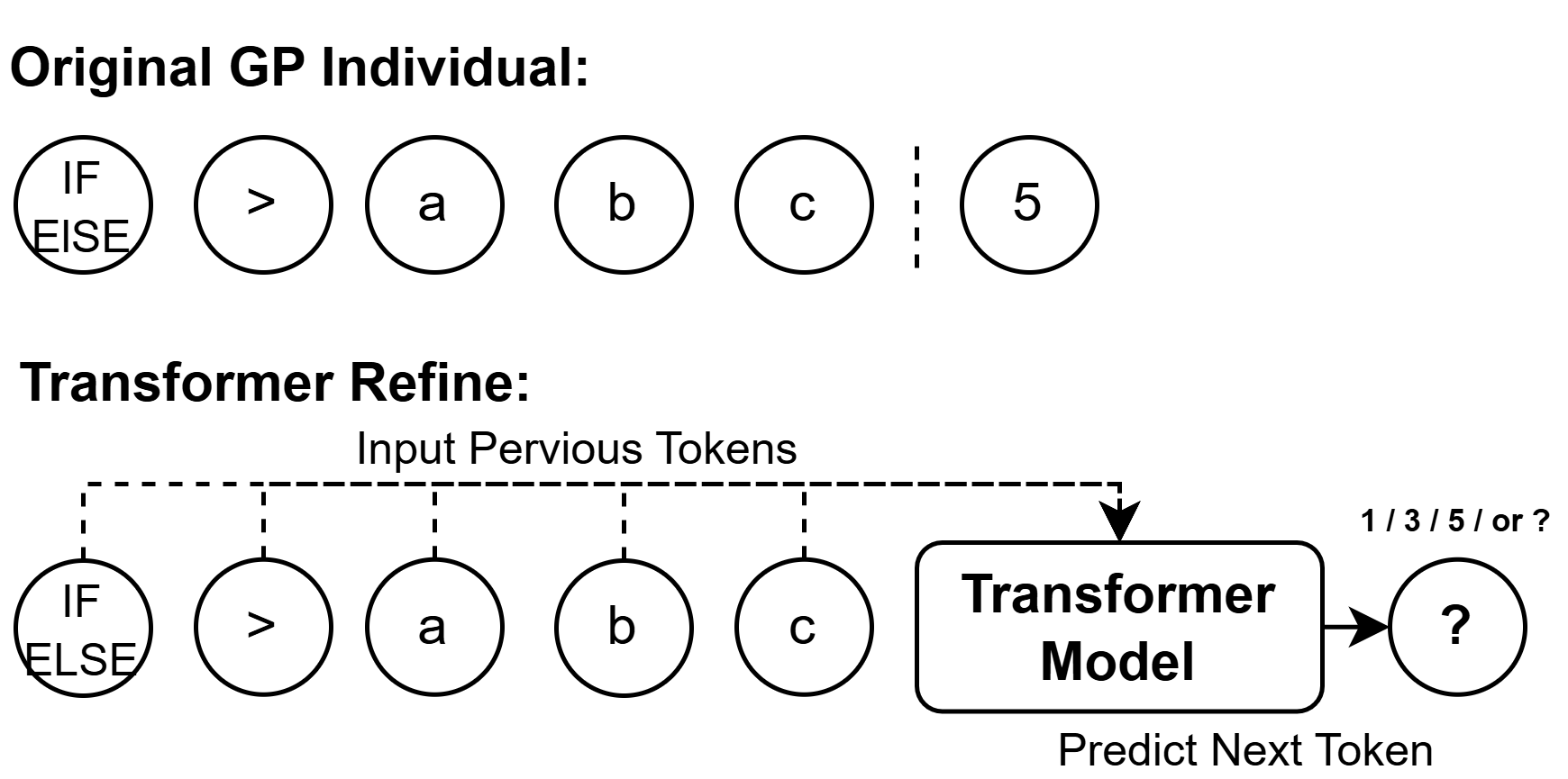}
\caption{Transformer Workflow}
\label{trans-str}
\end{figure}

As illustrated in Fig.~\ref{trans-str}, after GP generates an initial individual, the Transformer can refine it by leveraging its acquired knowledge to select better tokens, thereby optimizing performance. This predictive capability allows the Transformer to guide the search process more effectively, focusing on promising areas of the solution space and reducing reliance on random exploration. Consequently, the Transformer compensates for GP's limitations by enhancing the efficiency and effectiveness of the search for better scheduling schemes.

\begin{figure*}[htbp]
\centering
\includegraphics[width=15cm]{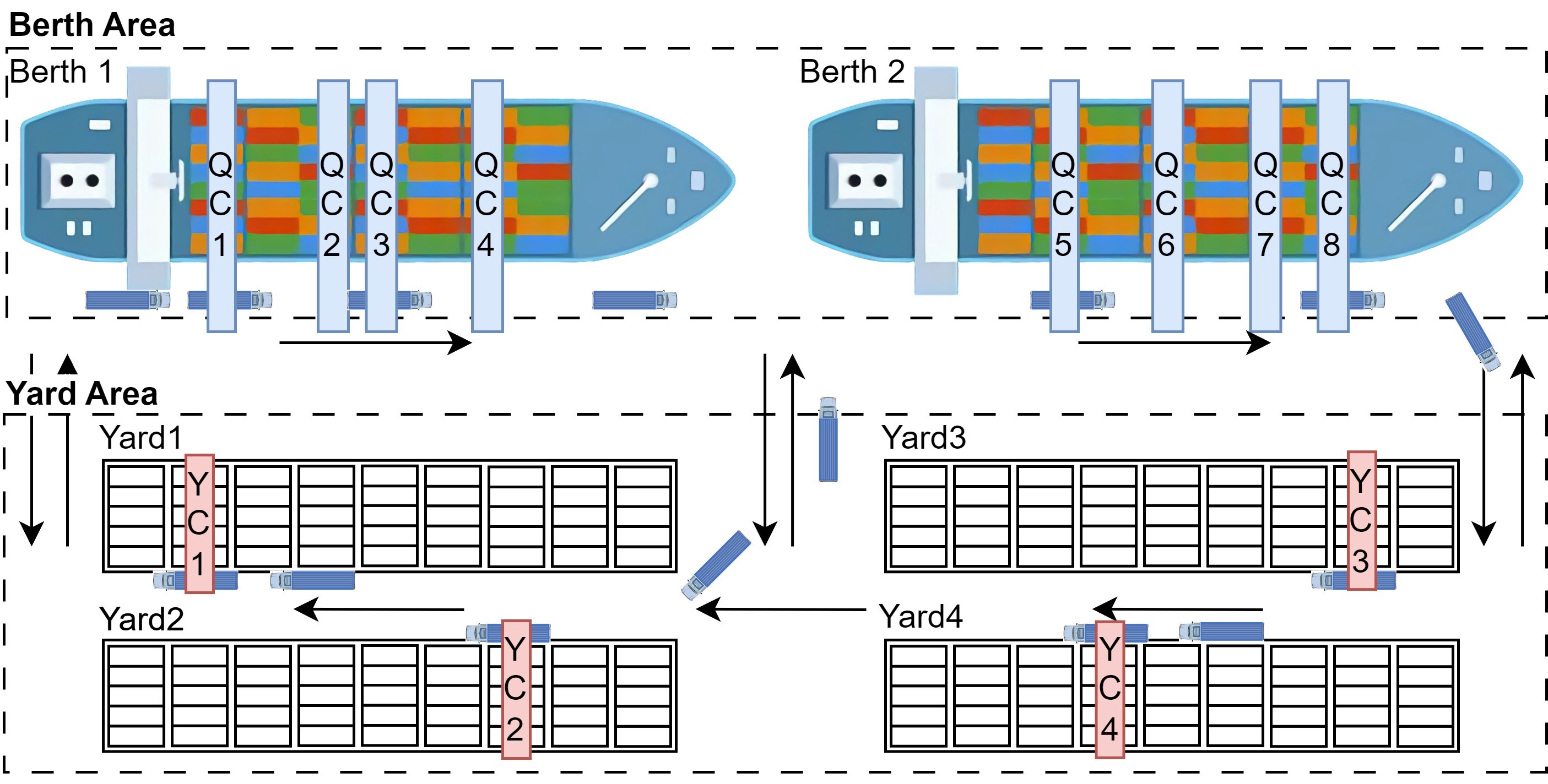}
\caption{A Sample Map of a Typical Container Port}
\label{port-map}
\end{figure*}

\section{Problem Statement}\label{section:problem}
Dynamic scheduling differs fundamentally from static scheduling due to the necessity of managing real-time uncertainties and continuous environmental changes. Experimental simulations often fail to capture the full complexity and unpredictability of real-world scenarios, leading to algorithms that perform well in controlled settings but struggle in practical applications. To bridge this gap, we have selected the container port truck routing problem as the focus of our study. This problem embodies the challenges of dynamic scheduling in a complex, real-world environment. Container port operations involve numerous unpredictable factors, such as variable weather conditions, interactions with external trucks not under terminal control, fluctuating cargo volumes, and human elements like individual driver behaviors. These factors introduce a high degree of uncertainty and complexity representative of many real-world dynamic scheduling problems.

By addressing the container port truck routing problem, we aim to test our algorithm in a setting that is both practically significant and inherently challenging. Successfully tackling this problem can demonstrate our algorithm's ability to handle complex uncertainties and adapt to dynamic changes—capabilities essential for effective dynamic scheduling. Moreover, if an algorithm can effectively solve the dynamic truck scheduling problem in a container terminal—a highly unpredictable and complex environment—it is well-equipped to handle less complex problems such as Automated Guided Vehicle (AGV) routing in warehouses, job shop scheduling in factories, and resource allocation in hospitals etc., where conditions are generally more controllable. Therefore, the insights gained from our study are transferable to other dynamic scheduling tasks across various industries, including logistics planning, supply chain management, and adaptive manufacturing systems.

As illustrated in Fig.~\ref{port-map}, the container port truck routing problem involves scheduling trucks to coordinate the transport of containers between quay cranes (QCs) and yard cranes (YCs). QCs, located near the ships, are responsible for loading and unloading containers from vessels, while YCs handle stacking and retrieval within the terminal. The primary objective is to efficiently assign trucks to facilitate container movement between ships and storage yards, thereby enhancing operational efficiency by minimizing delays, reducing idle times, and optimizing resource utilization.

The problem can be formally delineated as follows. An abstract container terminal is depicted as a directed graph, denoted by $G = (A, C)$, where $C = Q \cup Y$ constitutes the nodes representing the work operation points for all tasks. The sets $Q$ and $Y$ encompass all QCs and YCs, respectively. The set $A$ consists of direct driving connections between distinct nodes. The truck depot, represented by $d$, is the point from which all trucks depart at the commencement of the operation and return upon completion of all tasks. The set $V = \{v_1, v_2, v_3, \dots, v_m\}$ signifies the collection of $m$ available trucks for allocation. A function $\tau(x, y)$ maps two disparate operation points, $x \in C$ and $y \in C$, to the time required to traverse from one point to the other, reflecting the actual terminal road network.
The work instruction list encompasses all $n$ transport tasks in $T = \{t_1, t_2, t_3, \dots, t_n\}$. The container size for each task $t_i$ is denoted by $size_i$. The source and destination nodes for a given $t_i$ are represented by $a_i$ and $b_i$, respectively, with $a_i, b_i \in C$. Based on the diverse types of source and destination nodes, $ty_i$ is defined as the type of task $i$. $ty_i = 1$ signifies an unloading task, while $ty_i = 0$ corresponds to a loading task.

Within our problem framework, tasks are confined to transportation journeys exclusively between QCs and YCs. Consequently, $a_i$ and $b_i$ pertain to distinct crane-type node sets, either QCs or YCs. The maximum difference in task serial numbers, denoted as $q$, indicates the acceptable swapping order of unloading tasks (in this paper, $q=3$, considering the practicalities). The start time of service for $t_i$ at its source node is represented by $s_i$, while its completion time at the destination node is symbolized by $e_i$, where $s_i\in S = \{s_1,s_2,s_3,\dots,s_n\}$ and $e_i\in E = \{e_1,e_2,e_3,\dots,e_n\}$. Since a crane is required to either load or unload the container at the beginning and end of a task, the parameters $d_i$ and $h_i$ depict the operating time of $t_i$ at the source and destination nodes, respectively, and their sum is $r_i$.
The operation times at QCs and YCs are assumed to be stochastic and extracted from historical data.

To model the problem formally, the assignments of tasks to trucks are defined by the following binary variable in \eqref{eq2}:

\begin{equation}
    \alpha_{ij}=\left\{
    \begin{array}{rcl}
    1 & & t_j~\text{is~assigned~to}~v_i\\
    0 & & \text{otherwise}
    \end{array}
    \right.
    \label{eq2}
\end{equation}

\par
The following auxiliary variable indicates whether $t_k$ is serviced immediately after task $t_j$ by truck $v_i$.

\begin{equation}
    \beta_{ijk}=\left\{
    \begin{array}{rcl}
    1 & & t_k~\text{is~served~right~after}~t_j~\text{by}~v_i\\
    0 & & \text{otherwise}
    \end{array}
    \right.
    \label{eq3}
\end{equation}

The order of tasks belonging to a crane $c_i\in C$ is described by \eqref{eq1}. 

\begin{equation}
    \gamma_{ijk}=\left\{
    \begin{array}{rcl}
    1 & & t_k~\text{is~followed~by}~t_j~\text{in}~c_i\\
    0 & & \text{otherwise.}
    \end{array}
    \right.
    \label{eq1}
\end{equation}

The primary objective in truck scheduling problems for container terminals involves enhancing the company's profitability by increasing turnover and minimizing the waiting time of ships. To evaluate the extent to which this objective is accomplished, various metrics can be employed. 
In this study, we focus on the objective of \emph{TEU per hour (TEU/h)}, which is a metric calculating the quantity of Twenty-foot Equivalent Units (TEUs) processed hourly by all Quay Cranes (QCs) in use. The TEU, a standardized measure for containerized cargo, corresponds to a twenty-foot container's capacity. Port companies widely adopt this metric as a key indicator for benchmarking their operational efficiency against competitors.
It is noteworthy that the TEU/h metric is analogous to the makespan employed in numerous scheduling problems when the task set remains constant. Consequently, our truck scheduling problem can be modeled as follows:

\begin{equation}
   \max(\frac{\sum_{i=1}^{n}size_i}{\max(E) - \min(S)})
    \label{eq4} 
\end{equation}

\begin{equation}
    \sum_{i=1}^m\alpha_{ij} = 1 ~~ \forall t_j \in T
    \label{eq5}
\end{equation}

\begin{equation}
    \sum_{i=1}^m\sum_{k=1}^n\beta_{ijk} \le 1 \forall t_j\in T
    \label{eq6}
\end{equation}

\begin{equation}
    \begin{aligned}
        &\sum_{j=l}^{n}\sum_{k=1}^{l}\gamma_{ijk} \le q \cdot y_i ~~  l \in [1,n]
    \end{aligned}
    \label{eq11}
\end{equation}

\begin{equation}
    \begin{aligned}
        s_i= \max \left\{
        \begin{aligned}
        &\sum_{j=1}^n\sum_{k=1}^m\beta_{kji}\cdot(\tau(b_j,a_i) + e_j) \\
        &\tau(d,a_i)\cdot(1-\sum_{j=1}^n\sum_{k=1}^m\beta_{kji})
        \end{aligned}
        \right.
    \end{aligned}
    \label{eq9}
\end{equation}

\begin{equation}
    \begin{aligned}
        e_i= \max \left\{
        \begin{aligned}
        &\max (s_i,\sum_{j=1}^n\sum_{k=1}^m\gamma_{kji}\cdot e_j) + \tau(a_i,b_i) + r_i\\
        &\sum_{j=1}^n\sum_{k=1}^m\gamma_{kji}\cdot e_j + d_i
        \end{aligned}
        \right.
    \end{aligned}
    \label{eq10}
\end{equation}

The objective delineated in \eqref{eq4} represents the average production rate per unit of time (hour), where $\max{E}$ and $\min{S}$ correspond to the completion time of the final task and the start time of the first task, respectively. The constraint articulated in \eqref{eq5} guarantees that each task is assigned exclusively to one truck. In contrast, the constraint in \eqref{eq6} ascertains that each task is succeeded by a maximum of one other task or none if it is the truck's final task.
For each crane, constraint \eqref{eq11} following container terminal transportation rules, ensure that tasks involving the same crane cannot commence until the preceding task is concluded, except for the unloading tasks in QCs where the operational sequence can be interchanged between $sn=3$ neighboring tasks. Constraints \eqref{eq9} and \eqref{eq10} calculate the start and end times of tasks, verifying that tasks initiate crane operation only after completing preceding task operations.

Scheduling trucks within maritime container terminals is a complex task classified as NP-hard because it can be reduced to the well-known Vehicle Routing Problem~\cite{golden2008vehicle}. This means that as the size of the problem grows, finding the optimal solution becomes computationally infeasible due to exponential increases in complexity. Previous studies have primarily used metaheuristic approaches to address this challenge, often assuming that the crane operation times $r_i$ for tasks in the set $T$ are deterministic and constant. However, the operational environment of container terminals is characterized by inherent uncertainties and variations in crane operation times, making precise predictions difficult. As a result, the most common strategy in real-world applications is to implement a dynamic scheduling system that routes trucks in real time, adapting to changing conditions. This dynamic approach will be discussed in the next section.

\section{Methodologies}\label{section:method}
In this section, we present the methodologies employed to address the dynamic truck scheduling problem in container ports. We begin by discussing the design of a dynamic truck scheduling system tailored for real-time operations in a complex port environment. Within this system, we introduce four proposed methods that can serve as the core dynamic scheduling algorithms, including the Genetic Programming with Reinforcement Learning Trained RNN (GPRR) as a contrast group. Each method is examined in terms of its design, implementation, and suitability for dynamic scheduling in uncertain environments. We discuss the advantages and limitations of each approach, highlighting how they address the challenges inherent in the dynamic truck scheduling problem. In particular, we explain why the GPRT method is expected to outperform the others by effectively combining the strengths of GP and RL within a Transformer architecture.

\subsection{Dynamic Truck Scheduling System}\label{section:dynamic}
The container terminal dynamic truck scheduling system is designed to interact closely with both the dynamic dispatch algorithm and the Terminal Operating System (TOS). Unlike static scheduling, which generates schedules for all tasks in advance, dynamic scheduling operates in real time. The system continuously engages with the port environment by assigning tasks to idle trucks as they become available and by monitoring key environmental parameters such as vehicle distribution, crane operation conditions, and vehicle queue statuses.

\begin{figure}[htbp]
    \centering
    \includegraphics[width=8.8cm]{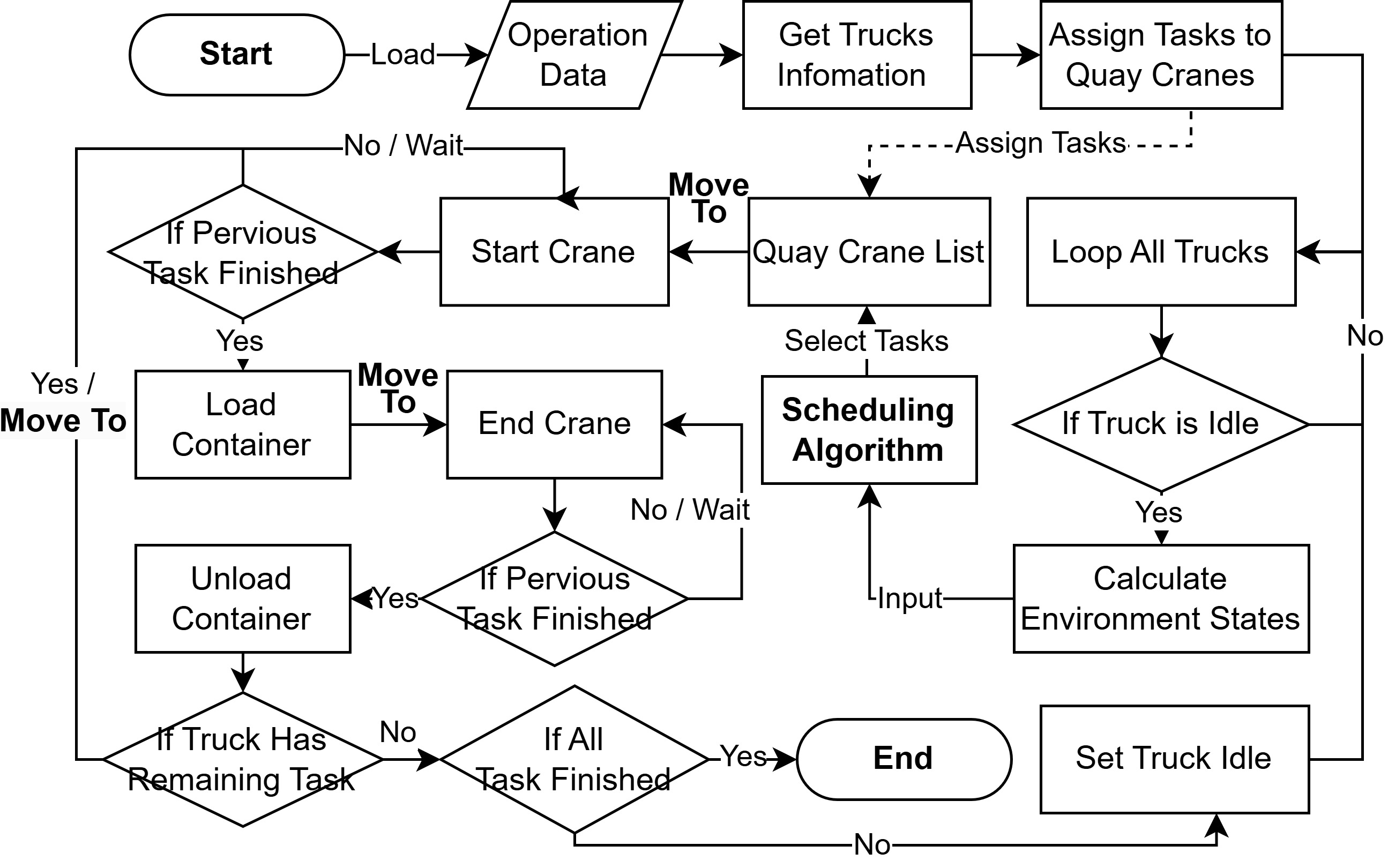}
    \caption{A Flowchart of Dynamic Truck Scheduling Systems}
    \label{dynamic}
\end{figure}

As illustrated in Fig.~\ref{dynamic}, the system follows a cyclical workflow where environmental information is updated, trucks are scheduled based on the latest data, and then the environmental information is updated again after scheduling. This continuous loop allows the system to adapt effectively to real-time changes in the port environment. By providing up-to-date information to the scheduling algorithms—including traditional heuristics like first-in-first-out and shortest transfer distance, as well as the proposed GP heuristics—the system can select the most appropriate tasks and support informed decision-making in dispatch operations.

In each cycle of this loop, if there are unfinished tasks remaining, the system schedules the most suitable task for an idle truck based on recommendations from the dispatch algorithm module. If all tasks have been completed, the system refrains from sending out further instructions and waits for new tasks to be assigned. This dynamic approach ensures that resources are utilized efficiently and the scheduling system can respond promptly to changes, thereby enhancing overall operational efficiency in the container terminal.

\subsection{Manual Heuristics}\label{section:manual}
Currently, many container terminals employ coordinators who manually optimize scheduling schemes and adjust the number of trucks assigned to each work queue. Despite relying on the coordinators' experience, most terminals manage to maintain relatively high operational efficiency, meeting market demands effectively. This success demonstrates that experienced operators have developed, over the years, practical knowledge and rules that significantly enhance truck scheduling at ports. To harness this expertise, we engaged with these operators through discussions, surveys, and questionnaires, encapsulating their insights into a manually crafted heuristic. We then applied this heuristic to real-world truck scheduling at marine container terminals, as outlined in Algorithm \ref{alg_manual}, serving as a baseline.

\begin{algorithm}
    \caption{Manually Crafted Heuristic Algorithms}
    \label{alg_manual}
    \begin{algorithmic}
    \Require Parameters $parameter$, Travel Time $t$
    \Function{$heuristic$}{$QC,truck$}
        \If{$crane\_truck\_num < desired\_trucks$}
        \State $score \leftarrow travel\_time * (truck\_num - prority)$
        \Else
        \State $score \leftarrow travel\_time * desired\_trucks$
        \EndIf
        \If{$truck\_num \ge truck\_limit$}
            \State $score \leftarrow score + 200000$
        \EndIf
        \State \Return $score$
    \EndFunction
    \end{algorithmic}
\end{algorithm}

This manually crafted heuristic algorithm utilizes several user-defined parameters based on the coordinators’ experience: \(desired\_trucks\) (the optimal number of trucks for a QC), \(priority\) (the importance of the QC), and \(truck\_limit\) (the maximum number of trucks per QC). It also integrates real-time observed variables: \(truck\_num\) (the current number of trucks servicing the QC) and \(travel\_time\) (the travel time from the current truck to the initial task’s source node for each QC). These factors are used to calculate a score for each available QC, reflecting the QC's operational preferences. The algorithm strategically schedules idle trucks to the QCs with the highest scores.

While this approach has been successfully implemented at our partner site, Ningbo-Zhoushan Port, its relatively simple structure and parameter set limit its overall performance. Additionally, it requires the expertise of specialists to reconfigure and adjust the parameters in response to changes in the operating scenario. Consequently, we propose the integration of machine learning techniques to automatically generate dispatch heuristics, aiming to enhance both performance and adaptability to scenario variations.

\subsection{GP with Logic Operators}

GP is a machine learning approach derived from the principles of biological evolution. It seeks to optimize a set of computer programs based on a predefined fitness function, which evaluates each program's performance in executing a specific computational task. The foundational premise of GP is the enhancement of problem solutions through continuous modifications of a population of individual solutions.

GP employs various representations, with the tree structure being the most prevalent and easily comprehensible one, which is also adopted in this study. Each GP entity is introduced into a port dispatch simulator, and the objective function, as shown in Equation \eqref{eq4}, is computed to determine the individual's fitness. Throughout the evolutionary procedure, GP adjusts each entity by implementing mutation and replication operations, ultimately selecting the individual demonstrating the highest fitness as the output.

\begin{figure}[!htbp]
\centering
\includegraphics[width=8.5cm]{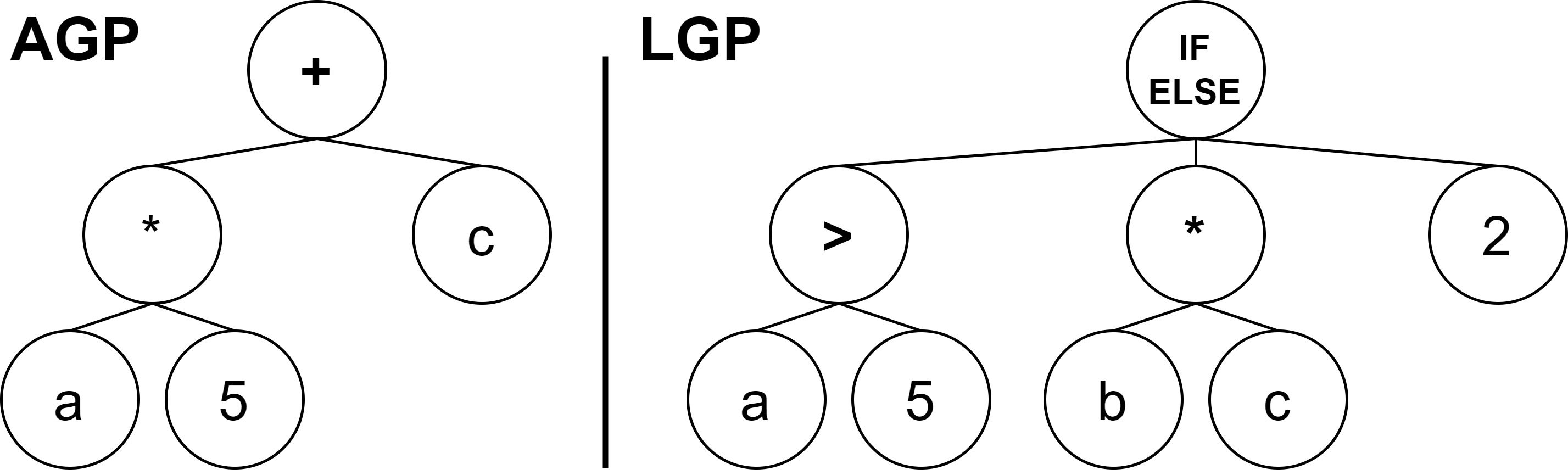}
\caption{AGP and LGP}
\label{AGP_LGP}
\end{figure}

As depicted in Fig. \ref{AGP_LGP}, we categorize GP methods into two types based on the inclusion of logical operators: Arithmetic Genetic Programming (AGP) and Logical Genetic Programming (LGP). Our previous study concluded that LGP outperforms AGP in the dynamic truck scheduling problem \cite{chen2022cooperative}. Therefore, this paper focuses exclusively on LGP, using the same operator configuration as in our previous publication \cite{chen2022cooperative}. This configuration includes arithmetic operators—addition, subtraction, multiplication, and protected division (where division by zero yields 1)—and logical operators such as greater than or equal to, less than or equal to, if-else, and, or, maximum, and minimum.

In the context of the dynamic truck scheduling problem, GP is employed to formulate heuristics for scheduling trucks. Similar to manual heuristics, environmental parameters are input into the GP heuristics to rank all tasks and select the optimal one, as described in the previous section. However, LGP suffers from unstable convergence, poor local search capabilities, and lacks knowledge learning. To address these issues and improve the algorithm's performance, we propose combining GP with RL.

\subsection{GP with RL Trained RNN}
While the heuristics or decision trees generated by GP offer excellent interpretability and have demonstrated strong performance in complex real-world dynamic scheduling problems, they still face several challenges during the search process. One significant issue is that the parameter settings within GP individuals can be redundant or suboptimal, and GP alone struggles to efficiently optimize these parameters. This limitation often requires extensive evolutionary computation to achieve local optimization, leading to increased computational effort. Additionally, the evolved GP individuals can become overly complex, which diminishes the interpretability of the heuristics—a key advantage of using GP in the first place.

To address these shortcomings, we propose combining GP with RL, specifically using Deep Reinforcement Learning (DRL) to train neural networks that optimize the GP heuristics. By integrating RL, we aim to enhance the learning capability of the system, allowing it to fine-tune the heuristics based on feedback from the environment, thus improving performance while maintaining interpretability.

With this foundation, the critical question becomes which type of neural network is most suitable for processing the heuristics generated by GP. As discussed earlier, GP individuals can be represented as sequences in bracket-free Polish notation, effectively transforming them into sequential data. RNN is a classic and well-established model for handling sequential data. Therefore, we first employ an RNN to integrate with GP, aiming to generate better scheduling schemes by leveraging the strengths of both methodologies.

By training the RNN using RL algorithms, the network can learn to generate and refine GP-like expressions that are more effective in solving dynamic scheduling problems. The RNN processes the sequential representation of GP individuals and predicts modifications that could improve performance. This approach allows the system to navigate the solution space more efficiently than random mutations in GP, as the RNN leverages learned knowledge to guide the search towards more promising candidates.

\subsubsection{RNN}
RNNs have been increasingly recognized for their ability to process sequential data, making them particularly suitable for producing heuristics in dynamic scheduling. By capturing temporal dependencies and patterns within sequences, RNNs can model the evolution of system states over time, which is crucial in dynamic scheduling, where decisions depend not only on the current state but also on past events and anticipated future changes. By integrating RNNs trained via RL into the GP framework, we aim to harness these strengths to generate more effective and adaptive scheduling heuristics. The RNN guides the search process by predicting promising candidate solutions based on learned temporal patterns, thereby improving upon the randomness of traditional GP mutations and crossovers.

\begin{figure}[!htbp]
\centering
\includegraphics[width=8.6cm]{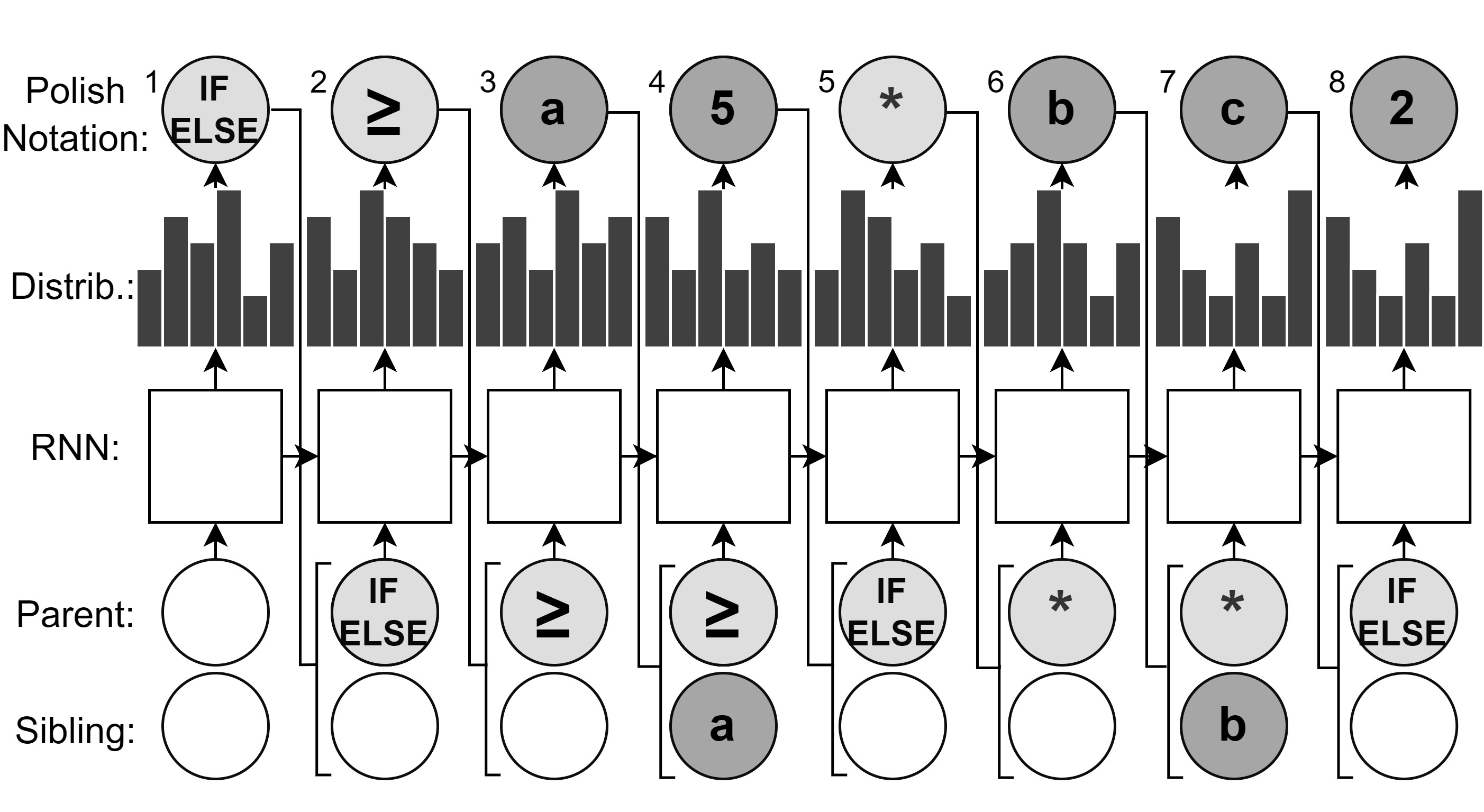}
\caption{The RNN Workflow}
\label{RNN_Flow}
\end{figure}

Like GP, RNN can also produce heuristics for truck schedules, but it employs Polish notation in this study. For instance, the LGP tree shown in Fig. \ref{AGP_LGP} corresponds to $[if\_else, \geq, a, 5, *, b, c, 2]$. As shown in Fig. \ref{RNN_Flow}, to generate this list, the RNN calculates probabilities for choosing different operators and parameters, considering both parent and sibling nodes. A token is then randomly selected based on these probabilities, leading to the stepwise construction of the Polish notation or heuristic.

Commonly, an RNN is selected such that the parameters $\theta$ render the likelihood of an expression tractable, thereby enabling the back-propagation of a differentiable loss function. For the $i$th token, denoted as $\tau_i$, its likelihood is conditionally independent given the preceding tokens $\tau_1, . . . , \tau_{(i-1)}$. Therefore, $p(\tau_i|\tau_{j \neq i}, \theta) = p(\tau_i|\tau_{j<i}, \theta)$. This study implements the RNN structure as informed by prior research \cite{chen2023neural}, specifically adopting an auto-regressive RNN constituted of a single-layer LSTM with 32 hidden nodes. 

After training, the RNN can randomly generate scheduling rules, effectively producing heuristics or Polish notation expressions similar to those generated by GP. This ability compensates for GP's reliance on random exploration within the population through evolutionary algorithms. By embedding learned knowledge into the neural network's weights, the RNN can make informed predictions based on different inputs, thereby enhancing the performance of the heuristics.

Furthermore, RNNs excel at capturing temporal dependencies and learning from sequential data, which allows them to acquire and transfer knowledge across different scenarios. This means the RNN can generalize from its training experience to adapt to new and varying conditions in dynamic scheduling environments. As a result, the integration of RNNs with GP not only improves the efficiency of the search process but also increases the adaptability and effectiveness of the scheduling heuristics.

However, effectively training the RNN to realize these advantages presents its own set of challenges. Determining the optimal training methodology and ensuring the RNN learns the most relevant features are critical steps. In the following section, we will discuss how to train the RNN to maximize its potential within this integrated framework.

\subsubsection{Reinforcement Learning}

Training RNNs traditionally relies on supervised learning, which requires known target outputs to compute a loss function. However, dynamic scheduling problems—which are NP-hard typically do not know the optimal scheduling solutions, especially during the planning phase when many uncertain events have yet to occur. This makes it challenging to determine the best possible scheduling scheme.

To address this issue, we employ RL to train the RNN. RL is suitable for scenarios where the optimal solution is unknown and must be discovered through environmental interaction. In RL, an agent learns to make decisions by receiving feedback in the form of rewards, aiming to maximize cumulative rewards over time. The RL framework comprises four main components: the environment, the state, the reward, and the policy.

In our GPRR approach, the RNN serves as the policy within the RL framework. The RNN generates heuristics that interact with the environment, effectively making actions. Unlike traditional RL agents that select actions directly based on the current state, our RNN does not interact with the environment directly. Instead, it interacts through the heuristics it generates, which are then used to make scheduling decisions.

By using RL to train the RNN, we enable it to learn from environmental feedback without requiring explicit knowledge of the optimal solutions. The RNN adjusts its internal parameters to maximize the expected reward, improving the quality of the heuristics over time. This approach allows the RNN to capture complex patterns and dependencies that are not easily discovered through random exploration or traditional training methods.

The details of RL are as follows:

\begin{itemize}
    \item \textbf{Environment:}
    The dynamic truck scheduling system described above serves as the training environment for the RL agent. By utilizing the map and historical data from our collaborator, the Ningbo Meishan Port, the system can simulate real-world port operations. It provides the current state ($S$) as input to the RL agent, and after the agent takes an action (assigning a truck), the system determines the subsequent state ($S'$) based on predefined rules and historical data. While the system is running, various performance metrics are calculated and used as rewards to assist in training the RL agent.
    \item \textbf{State:}
    The state, represented by a set of matrices, captures the current operating environment, enabling the DRL-GPHH and DRL-HH to choose appropriate actions based on varying conditions. In our study, the state is defined by several parameters reflecting the operational status of the trucks, tasks, queuing statuses of QCs and YCs, and the specific attributes of the QCs. The state matrix, with dimensions $i \times j$, includes $i$ parameters describing the status of each QC, with $j$ representing the number of QCs. These parameters are:
    \begin{itemize}
        \setlength{\itemsep}{0pt}
        \setlength{\parsep}{0pt}
        \setlength{\parskip}{0pt}
        \item Remaining and available task numbers for each QC.
        \item Number of trucks bound to each QC.
        \item Working status of the QC (0 for unload, 1 for load).
        \item QC type (0 for standard, 1 for remote control).
        \item Minimum truck travel time to the task's start crane.
        \item Minimum number of waiting trucks at the beginning and ending cranes.
        \item Average loading and unloading times for the beginning and ending cranes, respectively.
    \end{itemize}
    
    It is crucial to acknowledge the inherent uncertainty in this optimization problem of truck dispatching at container ports. Unlike typical state transitions modeled as $s' = E(s, a)$, here the transition includes an uncertainty parameter $u$, leading to $s' = E(s, a, u)$. 

    \item \textbf{Reword:}
    As shown in Equation \eqref{eq4}, the objective can only be calculated after all scheduling is complete, which leads to the issue of delayed rewards. Additionally, the large action space in reinforcement learning for dynamic scheduling presents challenges. These include problems with temporal credit assignment, where it's difficult to connect delayed rewards with the specific actions that caused them, potentially slowing the learning process or resulting in suboptimal strategies. Moreover, delayed rewards can disrupt the balance between exploration and exploitation, requiring extensive exploration to understand the long-term effects of actions, which might delay finding the optimal policy. Directly using the objective as the reward, as in methods like GP, can cause the reinforcement learning algorithm not to converge, making it unsuitable for use.
    
    To tackle these issues, we introduced a new reward function that merges reward shaping with imitation learning to better manage delayed rewards, thus enhancing credit assignment, facilitating efficient exploration, and promoting stable convergence. The formulated reward is expressed as \(reward = e_{i-1} - s_i - \delta \cdot cov(O_r, O_m)\), where \(\delta\) is a weight factor, \(cov\) is a covariance calculation, \(O_r\) is the task ranking by GPRR, and \(O_m\) is the manual heuristic ranking. The \(\delta\) parameter, set as \(\kappa/en\) (with \(\kappa\) being a scale factor and \(en\) the number of training episodes), adjusts the reward's sensitivity to similarity with manual heuristics. And \(\kappa\) is set to 10 in this study. This scaling allows the reward influence to decrease over training, initially leveraging expert heuristics for learning and then minimizing this reliance to avoid hampering the algorithm’s advancement toward optimal solutions. The component \(e_{i-1} - s_i\) is calculated post-task, addressing the delay issue, while the \(-\delta \cdot cov(O_r, O_m)\) component is immediate, ensuring timely feedback and guiding RL towards learning patterns akin to manual strategies.

    \item \textbf{Training Method:}
    There are numerous training methods for RL, each characterized by unique features. To streamline comparisons and reduce variability among these methods, this paper consistently employs the traditional Vanilla Policy Gradient (VPG) method \cite{williams1992simple} and follows all default settings. VPG utilizes the well-established REINFORCE rule, conducting training over the batch \(T\). The resulting loss function is defined as follows:
    \[
    L(\theta) = \frac{1}{|\tau|} \sum_{\tau \in T} (R(\tau) - b) \nabla_\theta \log(p(\tau|\theta))
    \]
    where \(b\) represents a baseline term or control variate, typically an exponentially weighted moving fitness average.

\end{itemize}

\subsubsection{GPRR Framework}

After figuring out GP, RNN, and RL, we combine them to form our GPRR framework, effectively integrating GP with RL. In this framework, the RNN is trained using both the heuristics it generates and those evolved by GP. Conversely, GP no longer needs to use traditional population initialization methods; instead, it can directly utilize the heuristics produced by the RNN as its initial population and continue to refine and train from there. By combining these two methods, the GPRR approach is expected to achieve superior performance in dynamic scheduling tasks.

\begin{figure}[!htbp]
\centering
\includegraphics[width=8.8cm]{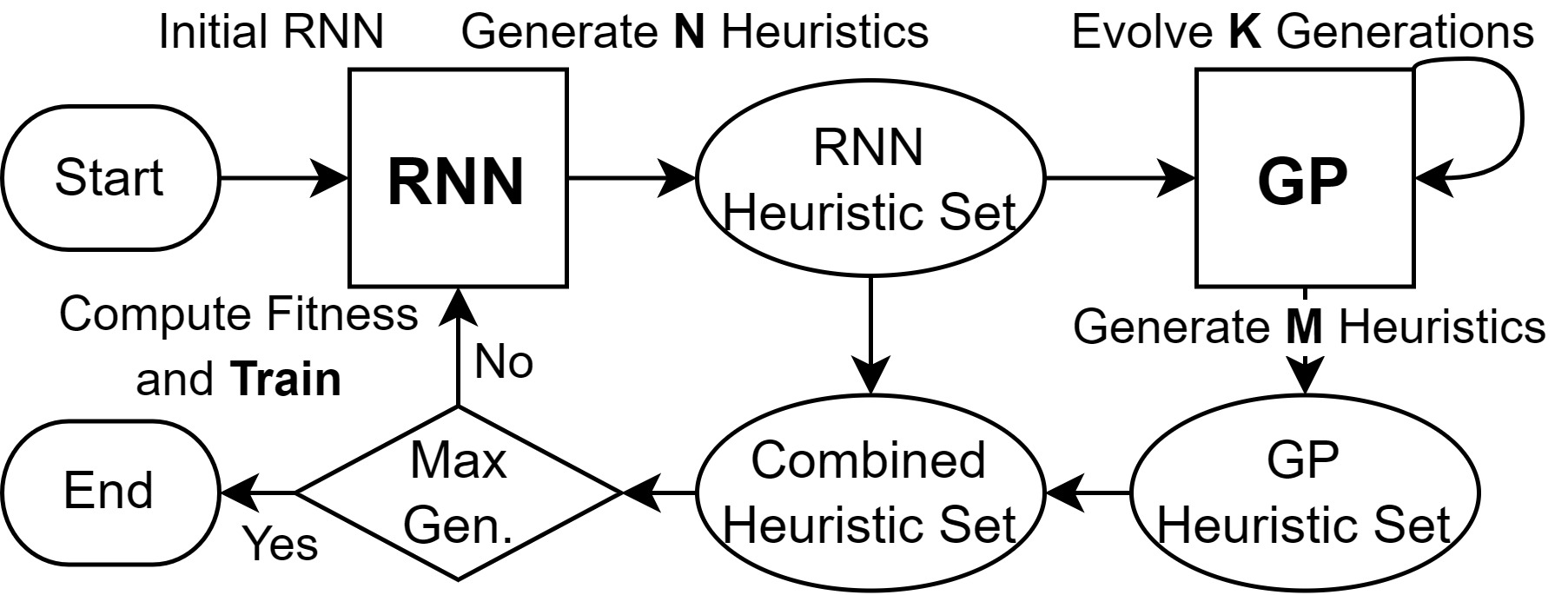}
\caption{The GPRR Framework}
\label{GPRR}
\end{figure}

The GPRRP model's framework, as depicted in Fig.~\ref{GPRR}, operates as follows: Initially, an RNN—configured with random neural networks, a combination of logical and arithmetic operators, and feature parameters—builds $N$ heuristics according to the process outlined in Fig.~\ref{RNN_Flow}. These heuristics serve as the initial population for subsequent GP processes. After $K$ generations, the GP produces $M$ heuristics. These are then merged with the initially generated $N$ RNN heuristics. If the maximum number of training generations is reached, the training concludes; otherwise, fitness values are calculated for each heuristic in this combined set. The RNN is then trained on these fitness scores and their corresponding heuristics, enabling it to refine the probability distribution of its outputs for diverse inputs and thus enhance heuristic generation. This cycle repeats with the RNN generating $N$ new heuristics for the subsequent iteration.

The GPRR model presents several key advantages that bolster both the efficacy of the training process and the quality of the solutions:

\begin{itemize}
\item \textbf{Population Diversity:} The integration of RNN enhances the diversity of the GP populations, promoting a more efficient evolutionary process. This enriched population encourages a wider exploration of the solution space, thus mitigating the risk of premature convergence to suboptimal solutions.

\item \textbf{Local Search Capabilities:} RNN augments GP with strong local search capabilities. By effectively exploring the solution landscape around promising areas identified by GP, RNN can uncover solutions that might be missed by GP's broader, population-based search.

\item \textbf{Training Efficiency:} GP supplies novel and high-quality training examples for RNN, thereby increasing the efficacy of RNN's training process. This reciprocal exchange of information allows both methodologies to continuously learn from each other, thereby enhancing the overall training efficiency.

\item \textbf{Complementarity:} GP and RNN complement each other, leading to an improvement in the overall quality of the solution. GP's aptitude for broad, population-based search is supplemented by RNN's expertise in refined local search, resulting in a comprehensive exploration and exploitation of the solution space.

\end{itemize}

In summary, the GPRR model harnesses the strengths of both GP and RNN, culminating in a robust, efficient, and versatile tool for addressing the dynamic truck dispatching problem.

\subsection{GP with RL Trained Transformer}

While the GPRR framework effectively combines GP with an RL-trained RNN to address the dynamic truck scheduling problem, it has inherent limitations due to the sequential nature of RNN. Specifically, RNN processes input tokens one at a time and primarily focuses on local context—such as parent nodes or nearby subtrees—when predicting the next token. This constraint limits their ability to capture long-range dependencies within heuristic expressions, which can hinder performance in complex dynamic scheduling problems.

To overcome these limitations, we propose replacing the RNN with a Transformer model, forming the GPRT framework. The Transformer architecture leverages self-attention mechanisms to consider the relationships between all tokens in a sequence simultaneously. This ability to model global dependencies allows the Transformer to capture intricate patterns and interactions within the heuristic expressions, leading to the generation of more effective heuristics.

\begin{figure}[!htbp]
\centering
\includegraphics[width=8.8cm]{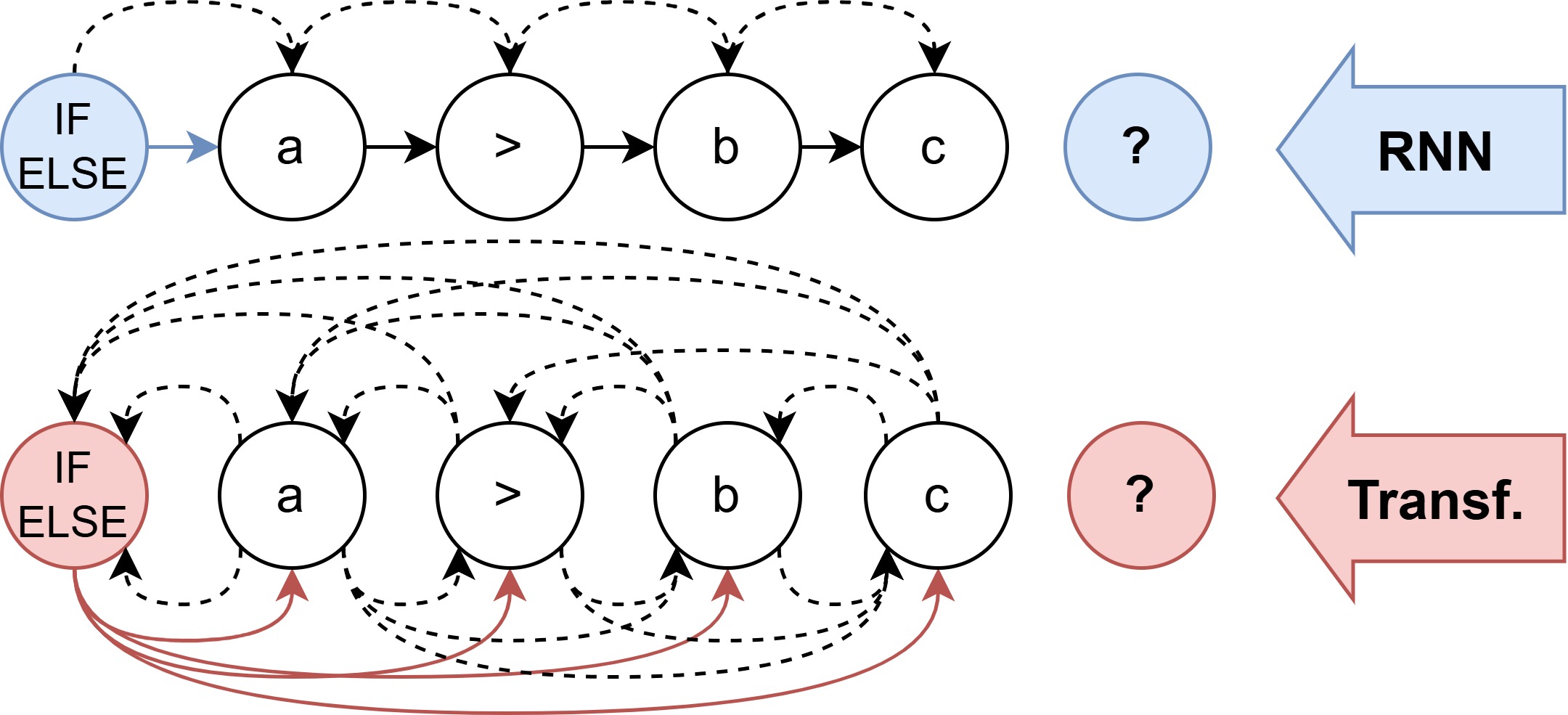}
\caption{RNN vs. Transformer}
\label{rnn_trans}
\end{figure}

As shown in Fig. \ref{rnn_trans}, by integrating GP with an RL-trained Transformer, we address the limitations of RNNs, which typically process information sequentially and make predictions based on that sequence. In contrast, the Transformer can independently attend to any token within the sequence, enabling it to make more accurate predictions. This capability significantly enhances the system's ability to produce high-quality scheduling heuristics.

The self-attention mechanism of the Transformer allows it to assess the importance of different parts of the heuristic during its generation. This leads to more informed and effective decision-making, as the model can focus on the most relevant information at each step of the process. We expect this approach to outperform previous models by more adeptly handling the complexities and uncertainties inherent in dynamic scheduling environments.

The GPRT replaces the neural network part in GPRR with a Transformer while keeping the rest the same:
\subsubsection{Transformer}

Transformers provide substantial improvements over traditional sequential models like RNNs by processing sequences in parallel, allowing efficient management of long-range dependencies. This capability is particularly critical for dynamic scheduling problems where such dependencies may span the entire sequence.

The architecture of the Transformer employed in this study focuses on the decoder component, which predicts subsequent tokens from an initial sequence based on conditional probabilities. The self-attention mechanism, central to this process, is described by the following equations:

\[
\text{Attention}(Q, K, V) = \text{softmax}\left(\frac{QK^T}{\sqrt{d_k}}\right)V
\]

Here, \(Q\), \(K\), and \(V\) are the query, key, and value matrices derived from the input tokens, respectively, and \(d_k\) represents the dimensionality of the keys. The output from the attention mechanism is utilized along with a feed-forward neural network within each decoder layer:

\[
\text{LayerOutput} = \text{FFN}(\text{AttentionOutput}) + \text{AttentionOutput}
\]

Each layer in the Transformer decoder includes residual connections and layer normalization to aid training and improve convergence:

\[
\text{LayerNorm}(x + \text{Sublayer}(x))
\]

where \(\text{Sublayer}(x)\) represents the function implemented by either the attention or feed-forward layers. The Transformer's ability to attend to all sequence elements simultaneously significantly enhances prediction quality for dynamic scheduling.

The Transformer structure utilized in this paper is adapted from our prior work \cite{chen2023transformer}, with modifications to its output layer to predict the next token rather than evaluating the fitness of previous tokens. This adaptation is key to its integration within the GPRT framework, where the Transformer's mechanisms are harnessed to generate scheduling heuristics. This approach aims to enhance both the accuracy and adaptability of these heuristics, particularly in complex environments, compared to traditional RNN-based systems.

The Transformer's capability to understand and process the entire list of tokens (heuristic) is crucial. It not only optimizes the performance of the heuristics but also potentially reduces their length and improves their comprehensibility. This ability to process comprehensive token sequences enables the Transformer to produce more efficient and understandable scheduling strategies, making it a superior choice for dynamic scheduling challenges.

\subsubsection{GPRT Framework}
To ensure a fair comparison between different algorithms, the GPRT framework differs from the GPRR model only by replacing the RNN in Fig.~\ref{GPRR} with a Transformer; all other components remain the same. This substitution allows us to isolate the impact of the Transformer's self-attention mechanism on the performance of the scheduling heuristics.

In the following sections, we proceed to the experimental part of our study to demonstrate the capabilities and superiority of GPRT in solving complex, real-world dynamic scheduling problems. Through comprehensive evaluations and comparisons, we aim to showcase how the integration of GP with an RL-trained Transformer enhances the system's ability to generate effective scheduling solutions, outperforming previous methods in handling the uncertainties and complexities inherent in dynamic scheduling environments.

\section{Experiments and Discussion}\label{section:experiment}

In this section, we conduct a comprehensive evaluation of the GPRR and GPRT frameworks, focusing on a real-world container port scheduling problem characterized by uncertain parameters. This analysis compares the integrated GP with RL-trained Transformer against baseline models including pure GP, RL, RNN, and standalone Transformer to highlight the advantages of our approach. The manual heuristic previously outlined in Section \ref{section:manual}, known for its robustness and effectiveness in practical port applications, serves as the benchmark for all comparative analyses (denoted as Improvement or Imp.). Additionally, this section includes ablation studies and sensitivity analyses to identify key factors contributing to the superior performance of GPRT.

\subsection{Experiment Design}

This research aims to develop an algorithm suitable for real-world port operations to enhance operational efficiency. All experimental data are derived from actual historical operating data of Ningbo Meishan Port, our collaborative partner. We utilized 20 days of operational data to create 10 training sets and 10 test sets, each containing 4,000 tasks. Experiments are carried out using an event-driven simulator we developed, with parameters adjusted according to historical data. The port setup includes two ship berths, 10 operating QC, and a variable number of container trucks ranging from 50 to 80.

Environmental parameters or GP/RNN/Transformer terminals/tokens follow the definitions from our previous research \cite{chen2022cooperative}. A total of 14 features are used to depict the current operational state of the port, including truck travel time, current number of QC trucks, number of waiting trucks, and number of remaining tasks. The GP configuration features a population size of 1024, with crossover, mutation, and reproduction rates set at 60\%, 30\%, and 10\%, respectively. The fitness function within the GP is denoted by equation \eqref{eq4}. All algorithms undergo 500 training generations, with the RNN's learning rate set at 0.001. Consistent with the GPRR and GPRT frameworks shown in Fig. \ref{GPRR} and following the methodology described in \cite{mundhenk2021symbolic}, K is set at 20, indicating the RNN's involvement every 20 GP training iterations. The size of the GP population, $M$, equals 1024, and $N$ is set at half of $M$, which is 512.

Each algorithm was executed 100 times using different random seeds to ensure robustness. We included the Cooperative Double-Layer Genetic Programming (CDGP) \cite{chen2022cooperative} and the Deep Reinforcement Learning Hyper-heuristic (DRL-HH) \cite{zhang2022deep} as control groups, representing advanced mono GP and RL models, respectively. The average results from these runs are presented in Tables \ref{result} for both the training and testing phases.

\subsection{Experimental Results}

In the training set of our experiment, the GPRT framework demonstrated superior performance, surpassing all other models, including GPRR, CDGP, and DRL-HH. The GPRT framework achieved a notable improvement of 18.77\% over the manual heuristics, showcasing its advanced ability to navigate the complexities and uncertainties inherent in dynamic scheduling environments. This significant enhancement is attributed to the Transformer’s capacity to analyze and integrate relationships across all tokens in the heuristic sequence effectively, ensuring that each decision is optimized based on a comprehensive understanding of the scheduling scenario. Moreover, the average TEU/h across all sets for GPRT was significantly higher than that for GPRR, which confirms the added benefit of replacing the RNN with a Transformer.

\begin{table}[htbp]
    \caption{Experimental Results of AGP, LGP, RNN and NN-GP (TEU/h)}
    \setlength\tabcolsep{2.5pt}
    \begin{center}
    \begin{threeparttable} 
    \begin{tabular}{cccccccc}
    \hline
    \textbf{Set} & \textbf{No.} & \textbf{Manual}  & \textbf{LGP} & \textbf{CDGP} & \textbf{DRL-HH} & \textbf{GPRR} & \textbf{GPRT} \\\hline
    
    \multirow{7}{*}{\textbf{Train}} 
    & 1 & 121.54 & 128.75 & 141.37 & 132.75 & 138.88 & 143.39 \\
    & 2 & 117.50 & 129.62 & 131.04 & 131.79 & 133.73 & 139.66 \\
    & 3 & 114.27 & 120.41 & 130.18 & 127.01 & 130.62 & 132.58 \\
    & 4 & 106.24 & 115.81 & 117.74 & 121.79 & 124.84 & 121.65 \\
    & 5 & 113.72 & 124.77 & 125.90 & 127.35 & 127.14 & 136.24 \\
     \cline{2-8}
    & \textbf{Avg.} & 114.65 & 123.87 & 129.25 & 128.14 & 131.04 & 136.17 \\
    & \textbf{Imp.} & \textbf{0.00\%} & \textbf{8.04\%} & \textbf{12.73\%} & \textbf{11.76\%} & \textbf{14.29\%} & \textbf{18.77\%} \\
     \hline
    
    \multirow{8}{*}{\textbf{Test}} 
     & 1 & 116.4 & 128.04 & 128.21 & 128.44 & 128.51 & 133.51\\
     & 2 & 115.63 & 123.96 & 128.06 & 131.09 & 129.39 & 134.84\\
     & 3 & 114.27 & 120.57 & 123.82 & 120.64 & 127.92 & 131.34\\
     & 4 & 106.01 & 111.08 & 113.70 & 116.10 & 115.70 & 122.89\\
     & 5 & 122.33 & 133.28 & 136.57 & 137.23 & 134.65 & 144.81\\
     \cline{2-8}
    & \textbf{Avg.} & 114.93 & 123.39 & 126.07 & 126.70 & 127.24 & 133.48\\
    & \textbf{Imp.} & \textbf{0.00\%} & \textbf{7.36\%} & \textbf{9.70\%} & \textbf{10.24\%} & \textbf{10.71\%} & \textbf{16.14\%}\\
\hline
    \end{tabular}
    \label{result}
    \begin{tablenotes}
        \footnotesize 
        \item[1] GPRR and GPRT differ from other algorithms,  $p < 0.05$.
      \end{tablenotes} 
    \end{threeparttable}
    \end{center}
\end{table}

In comparison, while CDGP and DRL-HH also showed improvements over manual heuristics, their performance enhancements were not as pronounced as those achieved by GPRR and GPRT. CDGP and DRL-HH displayed robust performance with a solid average increase over the manual heuristic. Yet, they still fell short of the capabilities demonstrated by the hybrid GPRR and GPRT model. This disparity highlights the limitations of traditional mono models like DRL-HH and CDGP in processing complex dependencies and dynamically adapting to changes—challenges that are effectively addressed by the combination of neural networks and evolutionary algorithms in GPRR and GPRT. Their key advantage lies in the ability to dynamically assess the importance of different aspects of the scheduling environment, leading to more informed and efficacious decision-making that significantly boosts operational efficiency.

Next, we put the trained CDGP,DRL-HH, GPRR and GPRT into a test environment completely different from the training environment with a broadly similar baseline. 

The experimental results compellingly demonstrate that the integration of GP and Reinforcement Learning RL within the GPRR and GPRT frameworks offers substantial advantages over traditional GP and RL methods. Interestingly, a notable discrepancy in performance between GPRR and GPRT during training highlights the challenges with generalization faced by RNN-based models, like those employed in GPRR. GPRT showed only a slight advantage over DRL-HH, improving by a mere 0.47\%, underscoring the limitations of RNNs in managing complex, dynamic scenarios where robust generalization across diverse operational conditions is crucial.

In contrast, GPRT, which utilizes a Transformer, did not suffer from significant performance fluctuations, demonstrating the robustness and consistency of Transformers in maintaining stable performance levels across both training and testing phases. This stability is particularly valuable in dynamic scheduling environments, where the ability to adapt to sudden changes without a loss in performance is crucial.

Despite a more substantial performance drop in GPRR during training, both GPRR and GPRT consistently outperformed conventional methods like LGP, CDGP, and DRL-HH in overall comparisons. This persistent superior performance indicates that the synergistic combination of GP and RL in these advanced frameworks effectively leverages the strengths of both approaches, leading to better outcomes even when individual components face challenges. The resilience and advanced capabilities of both GPRR and GPRT underscore their potential as powerful tools for dynamic scheduling, capable of surpassing traditional approaches.

In summary, the GPRT model effectively combines the capabilities of GP and the Transformer, exhibiting robust performance across both training and test scenarios. It proficiently manages the complexities of dynamic truck scheduling in multi-scenario ports, adeptly navigating inherent uncertainties. Given its successful application, the model demonstrates significant potential for extension to other dynamic scheduling problems, suggesting a broad scope for future applications.

\begin{table*}[htbp]
    \caption{Performance of Other Algorithms under Comparison (TEU/h)}
    \setlength\tabcolsep{5pt}
    \begin{center}
    \begin{tabular}{cccccccccccccc}
    \hline
    & \textbf{Perf.} & \textbf{Manual} & \textbf{Random} & \textbf{FIFO} & \textbf{STT} & \textbf{MTR} & \textbf{DQN} & \textbf{DDQN} & \textbf{PPO} & \textbf{RNN} & \textbf{Transformer} & \textbf{GPRR*} & \textbf{GPRT*} \\ \hline
        
        \multirow{2}{*}{\textbf{Train}} & \textbf{Avg.} 
        & 114.65 & 98.213 & 92.45 & 78.35 & 113.53 & 102.13 & 102.73 & 104.72 & 102.31 & 104.31 & 128.76 & 133.87 \\ 
        & \textbf{Imp.}
        & \textbf{N.A.} & \textbf{-14.34\%} & \textbf{-19.36\%} & \textbf{-31.66\%} & \textbf{-0.98\%} & \textbf{-10.92\%} & \textbf{-10.40\%} & \textbf{-8.66\%} & \textbf{-10.76\%} & \textbf{-9.02\%} & \textbf{12.31\%} & \textbf{16.76\%} \\ \hline
        
        \multirow{2}{*}{\textbf{Test}} & \textbf{Avg.} 
        & 114.93 & 99.32 & 92.86 & 76.23 & 113.92 & 101.324 & 101.66 & 104.39 & 101.38 & 103.98 & 126.54 & 132.18 \\ 
        & \textbf{Imp.}
        & \textbf{N.A.} & \textbf{-13.58\%} & \textbf{-19.20\%} & \textbf{-33.67\%} & \textbf{-0.88\%} & \textbf{-11.84\%} & \textbf{-11.55\%} & \textbf{-9.17\%} & \textbf{-11.79\%} & \textbf{-9.53\%} & \textbf{10.10\%} & \textbf{15.01\%} \\ \hline
    \end{tabular}
    \label{result_compare}
    \end{center}
\end{table*}

\subsection{Ablation Studies and Further Analysis}

In this section, we further investigate the performance of GPRR and GPRT through ablation and sensitivity analysis, with results presented in Table \ref{result_compare}. We began by introducing several traditional heuristic methods: random, first in first out (FIFO), shortest travel time (STT), and Most Tasks Remaining (MTR). Notably, both FIFO and STT performed worse than the manual heuristic, and in some cases, they were even outperformed by the random method. This is surprising, given that FIFO and STT are commonly used in traditional dynamic scheduling, highlighting the complexity of the problem as traditional methods struggle to address it. In contrast, the manual heuristic demonstrated performance comparable to MTR, indicating that despite the complexity of rules designed by human experts, there was no significant improvement over simpler heuristics.

Next, we compared GPRR and GPRT with three mainstream reinforcement learning (RL) methods: DQN, DDQN, and PPO, employing action and reward settings from our previous work \cite{chen2024deep}. It became evident that relying solely on these RL methods did not yield satisfactory results, as their performance lagged behind that of the manual heuristic. We then isolated the RNN and Transformer modules from GPRR and GPRT, finding their individual performances to be subpar. While the Transformer slightly outperformed the RNN, both were approximately 9\% worse than the manual heuristic. This suggests that standalone RL, RNN, and Transformer approaches struggle to effectively navigate the complex solution space of dynamic scheduling problems, leading to a lack of convergence and reduced local search capabilities.

Finally, to assess the impact of RNN and Transformer on GP population initialization, we introduced GPRR* and GPRT*, which are degraded versions of GPRR and GPRT without the RNN and Transformer for initialization. The results showed that GPRR* and GPRT* performed worse than their original counterparts. Interestingly, these versions also exhibited smaller performance declines in both the training and testing sets. This suggests that using RNN and Transformer for population initialization may reduce generalization capability, potentially affecting performance on the test set. We plan to explore this issue in greater depth in future research.

\begin{figure}[htbp]
\centering
\includegraphics[width=7cm]{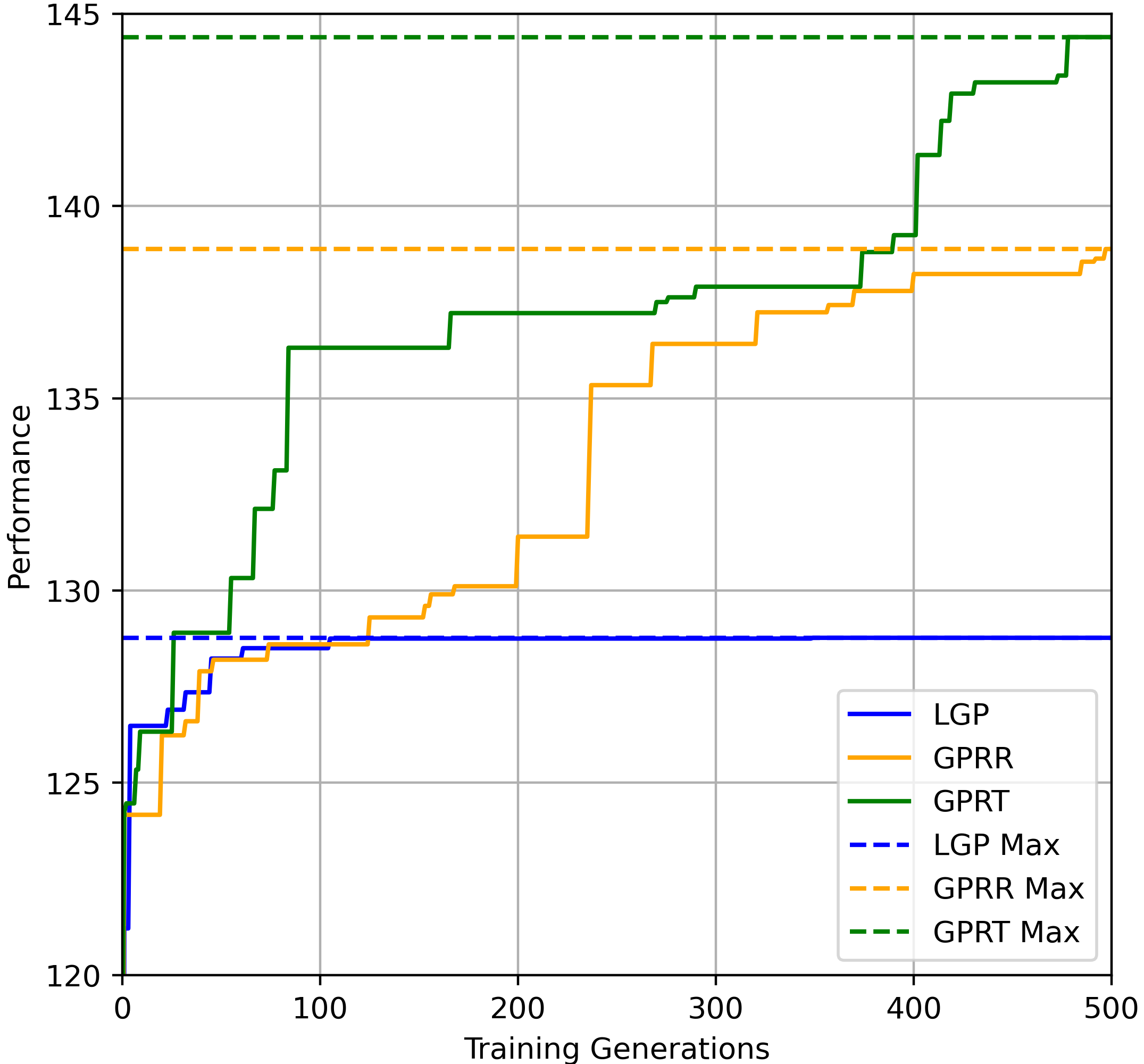}
\caption{The Performance of LGP, GPRR, GPRT on Training Set 1}
\label{train_res}
\end{figure}

Then, we present the training results for LGP, GPRR, and GPRT on Training Set 1, using the dataset that most closely aligns with the final average training performance in Fig. \ref{train_res}. It is clear that LGP converged early, showing no performance improvement after 100 generations. In contrast, both GPRR and GPRT demonstrated continuous performance enhancement throughout the training process. This supports our initial hypothesis that integrating RL-trained RNN and Transformer with GP can significantly improve GP's search capabilities, enabling the discovery of more effective dynamic scheduling rules. Furthermore, the relatively poor performance of RNN and Transformer, when used independently, emphasizes the importance of their combination in achieving superior results.

\begin{table}[htbp]
    \caption{The Average Token Number of LGP, RNN, Transformer, GPRR and GPRT}
    \begin{center}
    \begin{tabular}{cccccccc}
    \hline
     & \textbf{LGP} & \textbf{CDGP} & \textbf{RNN} & \textbf{Transformer} & \textbf{GPRR} &  \textbf{GPRT}\\\hline
    \textbf{Avg.} & 212.45  & 378.97  & 34.83  & 52.21  & 71.49  & 95.98 \\ 
\hline
    \end{tabular}
    \label{token_num}
    \end{center}
\end{table}

Moreover, we found that the results generated by the RNN and Transformer exhibited a smaller individual size (measured by token count, as shown in Table \ref{token_num}). We recorded the average token numbers after training LGP, CDGP, RNN, Transformer, GPRR, and GPRT for 500 generations. Notably, the methods utilizing RNN and Transformer produced fewer tokens, indicating smaller individual sizes and enhanced interpretability. Coupled with the impressive performance of GPRR and GPRT, this suggests that our approach combining GP and RL not only improves algorithm performance but also helps prune unnecessary components when processing GP-generated individuals. This indirectly enhances the effectiveness of GP evolution and final performance. The specific mechanisms by which RNN, Transformer, and GP collaborate in this evolutionary process remain an intriguing area for further exploration.

\section{Conclusion}\label{section:conclusion}

In this paper, we presented a novel approach that combines GP with RL through the frameworks of GPRR and GPRT, specifically designed for dynamic scheduling in complex environments. The experimental results demonstrated that our hybrid models consistently outperformed traditional methods, including standalone GP and RL approaches. By leveraging the strengths of both GP and RL, we achieved superior performance in identifying effective scheduling rules, showcasing the viability of this integrated methodology.

Our findings highlight the significant advantages of using RNNs and Transformers in conjunction with GP, as they not only enhanced the search capabilities of the algorithms but also contributed to the reduction of individual size, improving interpretability. The continuous performance improvements observed in GPRR and GPRT during training, in contrast to the early convergence of traditional methods like LGP, further validate our hypothesis that RL training can effectively optimize GP heuristics for dynamic scheduling tasks.

Looking forward, the successful application of this hybrid approach opens new avenues for research. Future work could focus on further refining the collaboration between RNNs, Transformers, and GP to enhance generalization capabilities and explore their applicability to other optimization problems within the transportation domain and beyond. By addressing the challenges identified in this study, such as the balance between complexity and interpretability, we aim to advance the field of dynamic scheduling and contribute to the development of more robust optimization frameworks.

\bibliographystyle{elsarticle-num}
\bibliography{ref}

\begin{thebibliography}{10}
\expandafter\ifx\csname url\endcsname\relax
  \def\url#1{\texttt{#1}}\fi
\expandafter\ifx\csname urlprefix\endcsname\relax\def\urlprefix{URL }\fi
\expandafter\ifx\csname href\endcsname\relax
  \def\href#1#2{#2} \def\path#1{#1}\fi

\bibitem{ouelhadj2009survey}
D.~Ouelhadj, S.~Petrovic, A survey of dynamic scheduling in manufacturing systems, Journal of scheduling 12 (2009) 417--431.

\bibitem{bukkur2018review}
K.~Bukkur, M.~Shukri, O.~M. Elmardi, A review for dynamic scheduling in manufacturing, The Global Journal of Researches in Engineering 18~(5-J) (2018) 25--37.

\bibitem{edelkamp2011heuristic}
S.~Edelkamp, S.~Schr{\"o}dl, Heuristic search: theory and applications, Elsevier, 2011.

\bibitem{schumann2022scheduling}
J.~M. Schumann, R.~Engelbeck, Scheduling: theory, algorithms, and systems (2022).

\bibitem{qiao2018novel}
F.~Qiao, Y.~Ma, M.~Zhou, Q.~Wu, A novel rescheduling method for dynamic semiconductor manufacturing systems, IEEE Transactions on Systems, Man, and Cybernetics: Systems 50~(5) (2018) 1679--1689.

\bibitem{begg2014uncertainty}
S.~H. Begg, M.~B. Welsh, R.~B. Bratvold, Uncertainty vs. variability: What’s the difference and why is it important?, in: SPE hydrocarbon economics and evaluation symposium, SPE, 2014, p. D011S003R002.

\bibitem{lu2017hybrid}
C.~Lu, L.~Gao, X.~Li, S.~Xiao, A hybrid multi-objective grey wolf optimizer for dynamic scheduling in a real-world welding industry, Engineering Applications of Artificial Intelligence 57 (2017) 61--79.

\bibitem{dixit2023contemporary}
A.~Dixit, S.~Jain, Contemporary approaches to analyze non-stationary time-series: Some solutions and challenges, Recent Advances in Computer Science and Communications (Formerly: Recent Patents on Computer Science) 16~(2) (2023) 61--80.

\bibitem{maxwell2017designing}
S.~E. Maxwell, H.~D. Delaney, K.~Kelley, Designing experiments and analyzing data: A model comparison perspective, Routledge, 2017.

\bibitem{walker2003defining}
W.~E. Walker, P.~Harremo{\"e}s, J.~Rotmans, J.~P. Van Der~Sluijs, M.~B. Van~Asselt, P.~Janssen, M.~P. Krayer~von Krauss, Defining uncertainty: a conceptual basis for uncertainty management in model-based decision support, Integrated assessment 4~(1) (2003) 5--17.

\bibitem{jin2005evolutionary}
Y.~Jin, J.~Branke, Evolutionary optimization in uncertain environments-a survey, IEEE Transactions on evolutionary computation 9~(3) (2005) 303--317.

\bibitem{fatemi2023scheduling}
S.~Fatemi-Anaraki, R.~Tavakkoli-Moghaddam, M.~Foumani, B.~Vahedi-Nouri, Scheduling of multi-robot job shop systems in dynamic environments: mixed-integer linear programming and constraint programming approaches, Omega 115 (2023) 102770.

\bibitem{deng2018optimal}
Z.~Deng, M.~Liu, H.~Chen, W.~Lu, P.~Dong, Optimal scheduling of active distribution networks with limited switching operations using mixed-integer dynamic optimization, IEEE Transactions on Smart Grid 10~(4) (2018) 4221--4234.

\bibitem{thilagavathi2014survey}
D.~Thilagavathi, A.~S. Thanamani, A survey on dynamic job scheduling in grid environment based on heuristic algorithms, arXiv preprint arXiv:1402.5205 (2014).

\bibitem{zhu2020making}
T.~Zhu, W.~Luo, C.~Bu, H.~Ning, Making use of observable parameters in evolutionary dynamic optimization, Information Sciences 512 (2020) 708--725.

\bibitem{firme2023agent}
B.~Firme, J.~Figueiredo, J.~M. Sousa, S.~M. Vieira, Agent-based hybrid tabu-search heuristic for dynamic scheduling, Engineering Applications of Artificial Intelligence 126 (2023) 107146.

\bibitem{chou2008simulated}
F.-D. Chou, H.-M. Wang, P.-C. Chang, A simulated annealing approach with probability matrix for semiconductor dynamic scheduling problem, Expert Systems with Applications 35~(4) (2008) 1889--1898.

\bibitem{renke2021review}
L.~Renke, R.~Piplani, C.~Toro, A review of dynamic scheduling: context, techniques and prospects, Implementing Industry 4.0: The Model Factory as the Key Enabler for the Future of Manufacturing (2021) 229--258.

\bibitem{xu2024genetic}
M.~Xu, Y.~Mei, F.~Zhang, M.~Zhang, Genetic programming and reinforcement learning on learning heuristics for dynamic scheduling: A preliminary comparison, IEEE Computational Intelligence Magazine 19~(2) (2024) 18--33.

\bibitem{shyalika2020reinforcement}
C.~Shyalika, T.~Silva, A.~Karunananda, Reinforcement learning in dynamic task scheduling: A review, SN Computer Science 1~(6) (2020) 306.

\bibitem{zhang2021genetic}
F.~Zhang, S.~Nguyen, Y.~Mei, M.~Zhang, Genetic Programming for Production Scheduling, Springer, 2021.

\bibitem{koza1994genetic}
J.~R. Koza, Genetic programming as a means for programming computers by natural selection, Statistics and computing 4 (1994) 87--112.

\bibitem{vanneschi2014survey}
L.~Vanneschi, M.~Castelli, S.~Silva, A survey of semantic methods in genetic programming, Genetic Programming and Evolvable Machines 15 (2014) 195--214.

\bibitem{medsker1999recurrent}
L.~Medsker, L.~C. Jain, Recurrent neural networks: design and applications, CRC press, 1999.

\bibitem{hochreiter1997long}
S.~Hochreiter, Long short-term memory, Neural Computation MIT-Press (1997).

\bibitem{vaswani2017attention}
A.~Vaswani, Attention is all you need, Advances in Neural Information Processing Systems (2017).

\bibitem{ahvanooey2019survey}
M.~T. Ahvanooey, Q.~Li, M.~Wu, S.~Wang, A survey of genetic programming and its applications, KSII Transactions on Internet and Information Systems (TIIS) 13~(4) (2019) 1765--1794.

\bibitem{tay2008evolving}
J.~C. Tay, N.~B. Ho, Evolving dispatching rules using genetic programming for solving multi-objective flexible job-shop problems, Computers \& Industrial Engineering 54~(3) (2008) 453--473.

\bibitem{banzhaf2024combinatorics}
W.~Banzhaf, T.~Hu, G.~Ochoa, How the combinatorics of neutral spaces leads genetic programming to discover simple solutions, in: Genetic Programming Theory and Practice XX, Springer, 2024, pp. 65--86.

\bibitem{riolo2010genetic}
R.~Riolo, T.~McConaghy, E.~Vladislavleva, Genetic programming theory and practice VIII, Vol.~8, Springer Science \& Business Media, 2010.

\bibitem{jakobovic2006dynamic}
D.~Jakobovi{\'c}, L.~Budin, Dynamic scheduling with genetic programming, in: European Conference on Genetic Programming, Springer, 2006, pp. 73--84.

\bibitem{chen2020data}
X.~Chen, R.~Bai, R.~Qu, H.~Dong, J.~Chen, A data-driven genetic programming heuristic for real-world dynamic seaport container terminal truck dispatching, in: 2020 IEEE Congress on Evolutionary Computation (CEC), Ieee, 2020, pp. 1--8.

\bibitem{bi2020genetic}
Y.~Bi, B.~Xue, M.~Zhang, Genetic programming with image-related operators and a flexible program structure for feature learning in image classification, IEEE Transactions on Evolutionary Computation 25~(1) (2020) 87--101.

\bibitem{chen2018improving}
Q.~Chen, B.~Xue, M.~Zhang, Improving generalization of genetic programming for symbolic regression with angle-driven geometric semantic operators, IEEE Transactions on Evolutionary Computation 23~(3) (2018) 488--502.

\bibitem{niazkar2023machine}
M.~Niazkar, M.~R. Goodarzi, A.~Fatehifar, M.~J. Abedi, Machine learning-based downscaling: Application of multi-gene genetic programming for downscaling daily temperature at dogonbadan, iran, under cmip6 scenarios, Theoretical and Applied Climatology 151~(1) (2023) 153--168.

\bibitem{sickel2010genetic}
K.~Sickel, J.~Hornegger, Genetic programming for expert systems, in: IEEE Congress on Evolutionary Computation, IEEE, 2010, pp. 1--8.

\bibitem{sette2001genetic}
S.~Sette, L.~Boullart, Genetic programming: principles and applications, Engineering applications of artificial intelligence 14~(6) (2001) 727--736.

\bibitem{winkler2007advanced}
S.~Winkler, M.~Affenzeller, S.~Wagner, Advanced genetic programming based machine learning, Journal of Mathematical Modelling and Algorithms 6~(3) (2007) 455--480.

\bibitem{kaelbling1996reinforcement}
L.~P. Kaelbling, M.~L. Littman, A.~W. Moore, Reinforcement learning: A survey, Journal of artificial intelligence research 4 (1996) 237--285.

\bibitem{li2017deep}
Y.~Li, Deep reinforcement learning: An overview, arXiv preprint arXiv:1701.07274 (2017).

\bibitem{silver2017mastering}
D.~Silver, J.~Schrittwieser, K.~Simonyan, I.~Antonoglou, A.~Huang, A.~Guez, T.~Hubert, L.~Baker, M.~Lai, A.~Bolton, et~al., Mastering the game of go without human knowledge, nature 550~(7676) (2017) 354--359.

\bibitem{levine2016end}
S.~Levine, C.~Finn, T.~Darrell, P.~Abbeel, End-to-end training of deep visuomotor policies, Journal of Machine Learning Research 17~(39) (2016) 1--40.

\bibitem{mao2016resource}
H.~Mao, M.~Alizadeh, I.~Menache, S.~Kandula, Resource management with deep reinforcement learning, in: Proceedings of the 15th ACM workshop on hot topics in networks, 2016, pp. 50--56.

\bibitem{liu2023dynamic}
C.-L. Liu, T.-H. Huang, Dynamic job-shop scheduling problems using graph neural network and deep reinforcement learning, IEEE Transactions on Systems, Man, and Cybernetics: Systems (2023).

\bibitem{wei2021recent}
H.~Wei, G.~Zheng, V.~Gayah, Z.~Li, Recent advances in reinforcement learning for traffic signal control: A survey of models and evaluation, ACM SIGKDD Explorations Newsletter 22~(2) (2021) 12--18.

\bibitem{jin2024container}
J.~Jin, T.~Cui, R.~Bai, R.~Qu, Container port truck dispatching optimization using real2sim based deep reinforcement learning, European Journal of Operational Research 315~(1) (2024) 161--175.

\bibitem{lin2022survey}
T.~Lin, Y.~Wang, X.~Liu, X.~Qiu, A survey of transformers, AI open 3 (2022) 111--132.

\bibitem{zeyer2019comparison}
A.~Zeyer, P.~Bahar, K.~Irie, R.~Schl{\"u}ter, H.~Ney, A comparison of transformer and lstm encoder decoder models for asr, in: 2019 IEEE Automatic Speech Recognition and Understanding Workshop (ASRU), IEEE, 2019, pp. 8--15.

\bibitem{kenton2019bert}
J.~D. M.-W.~C. Kenton, L.~K. Toutanova, Bert: Pre-training of deep bidirectional transformers for language understanding, in: Proceedings of naacL-HLT, Vol.~1, 2019, p.~2.

\bibitem{achiam2023gpt}
J.~Achiam, S.~Adler, S.~Agarwal, L.~Ahmad, I.~Akkaya, F.~L. Aleman, D.~Almeida, J.~Altenschmidt, S.~Altman, S.~Anadkat, et~al., Gpt-4 technical report, arXiv preprint arXiv:2303.08774 (2023).

\bibitem{kinnear1994advances}
K.~E. Kinnear, P.~J. Angeline, Advances in genetic programming, Vol.~3, MIT press Cambridge, MA, 1994.

\bibitem{golden2008vehicle}
B.~L. Golden, S.~Raghavan, E.~A. Wasil, The vehicle routing problem: latest advances and new challenges, Vol.~43, Springer Science \& Business Media, 2008.

\bibitem{chen2022cooperative}
X.~Chen, R.~Bai, R.~Qu, H.~Dong, Cooperative double-layer genetic programming hyper-heuristic for online container terminal truck dispatching, IEEE Transactions on Evolutionary Computation 27~(5) (2022) 1220--1234.

\bibitem{chen2023neural}
X.~Chen, F.~Bao, R.~Qu, J.~Dong, R.~Bai, Neural network assisted genetic programming in dynamic container port truck dispatching, in: 2023 IEEE 26th International Conference on Intelligent Transportation Systems (ITSC), IEEE, 2023, pp. 2246--2251.

\bibitem{williams1992simple}
R.~J. Williams, Simple statistical gradient-following algorithms for connectionist reinforcement learning, Machine learning 8 (1992) 229--256.

\bibitem{chen2023transformer}
X.~Chen, J.~Dong, R.~Qu, R.~Bai, Transformer surrogate genetic programming for dynamic container port truck dispatching, in: International Conference on Bio-Inspired Computing: Theories and Applications, Springer, 2023, pp. 276--290.

\bibitem{mundhenk2021symbolic}
T.~N. Mundhenk, M.~Landajuela, R.~Glatt, C.~P. Santiago, D.~M. Faissol, B.~K. Petersen, Symbolic regression via neural-guided genetic programming population seeding, arXiv preprint arXiv:2111.00053 (2021).

\bibitem{zhang2022deep}
Y.~Zhang, R.~Bai, R.~Qu, C.~Tu, J.~Jin, A deep reinforcement learning based hyper-heuristic for combinatorial optimisation with uncertainties, European Journal of Operational Research 300~(2) (2022) 418--427.

\bibitem{chen2024deep}
X.~Chen, R.~Bai, R.~Qu, J.~Dong, Y.~Jin, Deep reinforcement learning assisted genetic programming ensemble hyper-heuristics for dynamic scheduling of container port trucks, IEEE Transactions on Evolutionary Computation (2024).

\end{thebibliography}

\end{document}